\documentclass[lettersize,journal]{IEEEtran}
\usepackage{algorithmic}
\usepackage[linesnumbered,ruled,vlined]{algorithm2e}

\usepackage{amsmath,amsfonts}
\usepackage{amssymb,amsthm}
\usepackage{array}
\usepackage{textcomp}
\usepackage{stfloats}
\usepackage{url}
\usepackage{cite}
\usepackage{blindtext}
\usepackage{enumitem}
\usepackage{hyperref}
\usepackage{booktabs}
\usepackage{multicol}
\usepackage{multirow}
\usepackage{xcolor}
\usepackage{graphicx}
\usepackage{subcaption}
\usepackage{tikz}
\usepackage{pgfplots}
\usepgfplotslibrary{colorbrewer}
\usepgfplotslibrary{groupplots}
\pgfplotsset{width=10cm,compat=1.17}
\usepackage{pifont}
\newcommand{\cmark}{\ding{51}}%
\newcommand{\xmark}{\ding{55}}%

\theoremstyle{definition}

\newtheorem{example}{Example}
\newtheorem{Problem Definition}{Problem Definition}

\usepackage{titlesec}
\titlespacing\section{0pt}{3pt}{2pt}
\titlespacing\subsection{0pt}{3pt}{2pt}
\titlespacing\subsubsection{0pt}{3pt}{2pt}

\begin{document}

\title{FRAMU: Attention-based Machine Unlearning using Federated Reinforcement Learning}


\author{Thanveer Shaik*,  Xiaohui Tao*, Lin Li, Haoran Xie, Taotao Cai, Xiaofeng Zhu, and Qing Li
\thanks{*Corresponding authors: Thanveer Shaik (email: Thanveer.Shaik@usq.edu.au) and  Xiaohui Tao (email: Xiaohui.Tao@usq.edu.au) are with the School of Mathematics, Physics, and Computing, University of Southern Queensland, Queensland, Australia}
\thanks{Taotao Cai is with the School of Mathematics, Physics, and Computing, University of Southern Queensland, Queensland, Australia (e-mail: Taotao.Cai@usq.edu.au).}
\thanks{Lin Li is with the School of Computer and Artificial Intelligence, Wuhan University of Technology, China (e-mail: cathylilin@whut.edu.cn)}
\thanks{Haoran Xie is with the Department of Computing and Decision Sciences, Lingnan University, Tuen Mun, Hong Kong (e-mail: hrxie@ln.edu.hk)}
\thanks{Xiaofeng Zhu is with the University of Electronic Science and Technology of China (e-mail: seanzhuxf@gmail.com)}
\thanks{Qing Li is with the Department of Computing, Hong Kong Polytechnic University, Hong Kong Special Administrative Region of China (e-mail: qing-prof.li@polyu.edu.hk).}
}



\maketitle

\begin{abstract}

Machine Unlearning, a pivotal field addressing data privacy in machine learning, necessitates efficient methods for the removal of private or irrelevant data. In this context, significant challenges arise, particularly in maintaining privacy and ensuring model efficiency when managing outdated, private, and irrelevant data. Such data not only compromises model accuracy but also burdens computational efficiency in both learning and unlearning processes. To mitigate these challenges, we introduce a novel framework, Attention-based Machine Unlearning using Federated Reinforcement Learning (FRAMU). This framework incorporates adaptive learning mechanisms, privacy preservation techniques, and optimization strategies, making it a well-rounded solution for handling various data sources, either single-modality or multi-modality, while maintaining accuracy and privacy. FRAMU's strength lies in its adaptability to fluctuating data landscapes, its ability to unlearn outdated, private, or irrelevant data, and its support for continual model evolution without compromising privacy. Our experiments, conducted on both single-modality and multi-modality datasets, revealed that FRAMU significantly outperformed baseline models. Additional assessments of convergence behaviour and optimization strategies further validate the framework's utility in federated learning applications. Overall, FRAMU advances Machine Unlearning by offering a robust, privacy-preserving solution that optimizes model performance while also addressing key challenges in dynamic data environments.
\end{abstract}

\begin{IEEEkeywords}
Machine Unlearning, Privacy, Reinforcement Learning, Federated Learning, Attention Mechanism.
\end{IEEEkeywords}

\section{Introduction}

The widespread availability of decentralized and heterogeneous data sources has created a demand for Machine Learning models that can effectively leverage this data while preserving privacy and ensuring accuracy~\cite{kumar2021tp2sf}. Traditional approaches struggle to handle the continual influx of new data streams, and the accumulation of outdated or irrelevant information hinders their adaptability in dynamic data environments~\cite{nian2020review,wahab2021federated}. Moreover, the presence of sensitive or private data introduces concerns regarding data breaches and unauthorized access, necessitating the development of privacy-preserving techniques~\cite{al2019privacy}. The concept of the ``right to be forgotten'' allows individuals to have their personal information removed from online platforms, although there's no universal agreement on its definition or its status as a human right~\cite{politou2018forgetting}. Despite this, countries like Argentina, the Philippines, and large parts of the EU are working on regulations~\footnote{https://link.library.eui.eu/portal/The-Right-To-Be-Forgotten--A-Comparative-Study/tw0VHCyGcDc/}. Therefore, there is a  pressing need to advance the field of Machine Unlearning to ensure both adaptability and privacy in Machine Learning applications.


\begin{figure}
    \centering
    \includegraphics[scale=0.4]{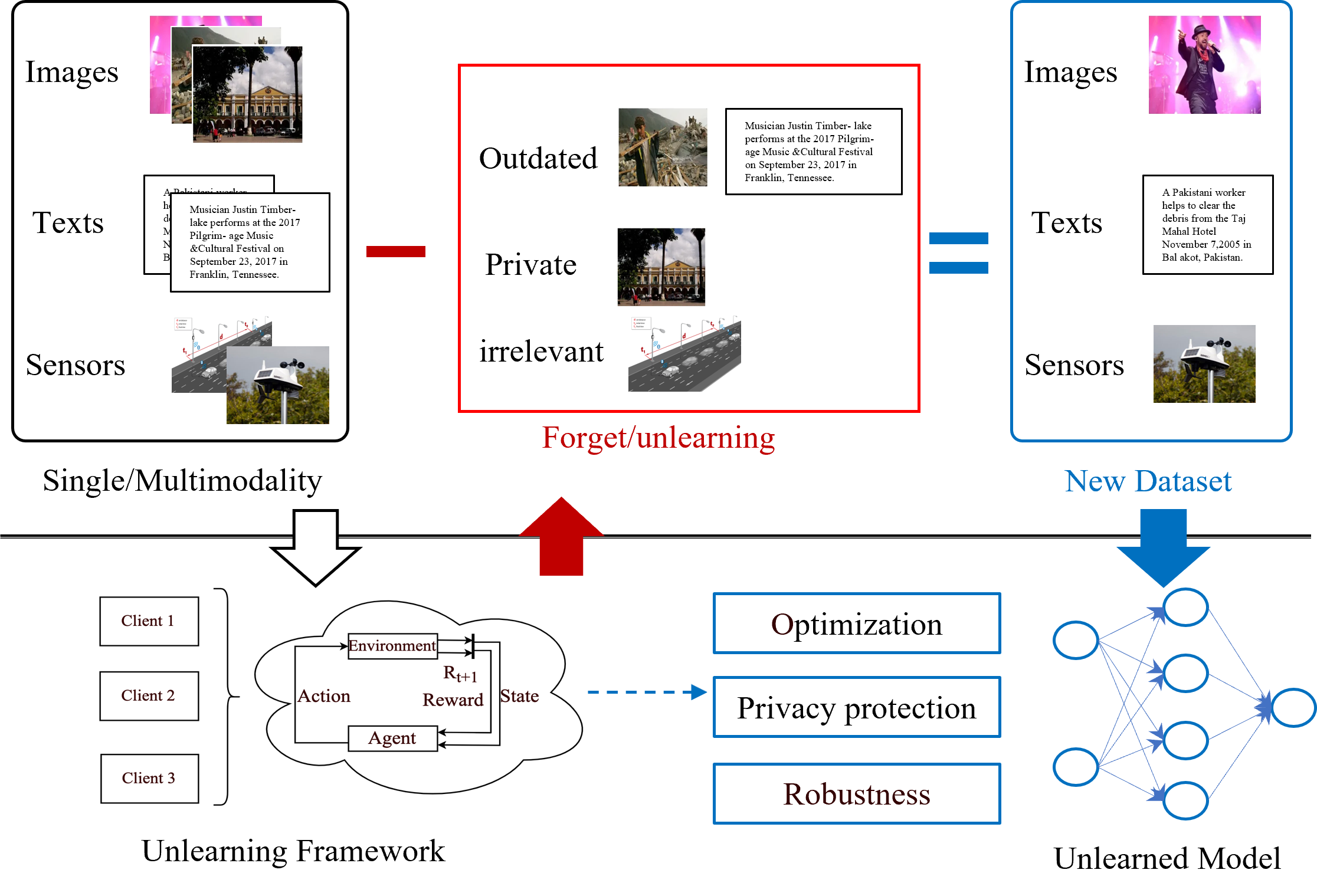}
    \caption{\small{Graphical abstract depicts the evolution of the FRAMU framework}}
    \label{fig:framu_abstract}
\end{figure}


\begin{example}\label{example_1}
\textit{In a landmark 2014 decision that underscored the pressing need for Machine Unlearning, a Spanish court ruled in favor of an individual who sought the removal of specific, outdated Google search results related to a long-settled debt~\cite{globocnik2020right}. This verdict not only led to Google taking down the search results but also influenced broader European Union policies on the subject, emphasizing the urgent need for mechanisms that can efficiently erase outdated or private information from Machine Learning models without sacrificing accuracy. This critical requirement for Machine Unlearning is further highlighted by high-profile cases such as that of James Gunn, the famed writer and director, who was dismissed by Disney in 2018 when old, inappropriate tweets resurfaced~\cite{fortner2020decision}. Although social media platforms like Facebook offer features like "Off-Facebook Activity" to disconnect user data from third-party services, this does not guarantee the complete erasure of that data from the internet~\footnote{https://www.facebook.com/help/2207256696182627}. Together, these instances accentuate the growing imperative for the development of robust Machine Unlearning technologies, especially in an era where data privacy regulations are continuously evolving and the "right to be forgotten" is increasingly recognized as essential.}
\end{example}

\textit{Challenges.} In today's digitally connected environment, data is distributed in various forms and from different sources, such as sensors, text documents, images, and time series data. For unlearning outdated or private data, Machine Unlearning presents unique challenges depending on whether it is a single type of data (known as single-modality) or multiple types of data (referred to as multimodality)~\cite{vasilevski2023meta}. With single-modality data, the issue primarily lies in the build-up of outdated or irrelevant information, which can negatively affect the model's effectiveness and precision~\cite{sun2023unified,malekloo2022machine}. On the other hand, multimodality situations are even more complicated. Here, each type of data can have different characteristics and varying contributions to the overall model's performance~\cite{ngiam2011multimodal,zhou2019review}. As we discussed in example~\ref{example_1}, the need to unlearn outdated or private data is most important. This ensures individuals have the ``right to be forgotten'' about their information in publicly available platforms. However, the unlearning needs to happen in both single-modality and multimodality data to make it a holistic unlearning.

Distributed learning systems, particularly federated learning, have made significant strides forward in enabling Machine Learning models to train on decentralized data, offering the dual advantage of reduced communication costs and enhanced privacy~\cite{fernandez2022privacy,nguyen2022federated}. Notable efforts have been made to incorporate Differential Privacy (DP) into these systems~\cite{li2021survey}, ensuring robust privacy safeguards through techniques like DP-SGD and DP-FedAvg~\cite{abadi2016deep, mcmahan2017learning}. However, these existing frameworks face limitations when confronted with the dynamic nature of data distribution, an intrinsic challenge in distributed learning~\cite{chen2021unified}. Although some efforts have been made in Machine Unlearning to address data irrelevancy over time, such as Sharded, Isolated, Sliced, and Aggregated(SISA) training methods, these solutions often operate in isolation from privacy-preserving mechanisms~\cite{bourtoule2021machine,jegorova2022survey}. This bifurcation leaves a crucial research gap: the absence of a unified approach that addresses both privacy concerns and the adaptability requirements in the face of ever-changing data landscapes. There is a need to bridge this gap by providing an integrated solution for robust privacy measures and efficient selective unlearning, thereby enabling Machine Learning models to be both secure and adaptable in dynamic, distributed environments.

The primary challenges in Machine Unlearning involve addressing the buildup of outdated or irrelevant information in single-modality data, which affects model precision, and handling the complexity of multimodality data where each type contributes differently to model performance. Additionally, current distributed learning systems, while advancing privacy and reducing communication costs, face limitations in adapting to dynamic data distributions and integrating robust privacy measures with efficient unlearning mechanisms, highlighting a need for a unified approach that ensures both security and adaptability in rapidly evolving data environments.

To address these challenges, we propose an Attention-based Machine Unlearning using Federated Reinforcement Learning (FRAMU) as shown in Fig.~\ref{fig:framu_abstract}. By integrating federated learning, adaptive learning mechanisms, and privacy preservation techniques, FRAMU aims to leverage the diverse and dynamic nature of data in both single-modality and multimodality scenarios, while upholding privacy regulations and optimizing the learning process. An attention mechanism is incorporated into FRAMU to ensure responsible and secure handling of sensitive information across modalities. FRAMU leverages reinforcement learning and adaptive learning mechanisms to enable models to dynamically adapt to changing data distributions and individual participant characteristics in both single-modality and multimodality scenarios. This adaptability facilitates ongoing model evolution and improvement in a privacy-preserving manner, accommodating the dynamic nature of the data present in federated learning scenarios. In addition to addressing the challenges associated with unlearning outdated, private, and irrelevant data in both single-modality and multimodality scenarios, FRAMU offers valuable insights into the convergence behaviour and optimization of the federated learning process. The major contributions of our work are as follows:

\begin{itemize}
    \item We propose an adaptive unlearning algorithm using an attention mechanism to adapt to changing data distributions and participant characteristics in single-modality and multimodality scenarios.
    \item We develop a novel design to personalize the unlearning process using the FedAvg mechanism~\cite{mcmahan2017communication} and unlearn the outdated, private, and irrelevant data.
    \item We propose an efficient unlearning algorithm that demonstrates fast convergence and achieves optimal solutions within a small number of communication rounds.
    \item We conduct extensive experiments to demonstrate the efficiency and effectiveness of the proposed approach using real-world datasets.
\end{itemize}

\textit{Organization.} In Section~\ref{relatedworks}, we review related works. Section~\ref{problem} outlines the problem addressed in this study. We present the proposed framework FRAMU in Section~\ref{methods}. The applications of FRAMU in single-modality and multimodality are discussed in Section~\ref{applicationsofframu}. In Section~\ref{experiment}, we present the experimental setup and the evaluation results of the proposed framework, along with convergence and optimization analysis. Section~\ref{discussion} delves into the implications of the proposed framework. Finally, in Section~\ref{conclusion}, we conclude the paper.

\section{Related Works}\label{relatedworks}
The importance of data privacy in distributed learning systems has garnered significant attention, especially when handling sensitive types of data like medical or behavioral information~\cite{wu2020privacy}. Differential Privacy (DP), a mathematically rigorous framework for ensuring individual privacy, has been widely adopted for this purpose~\cite{liang2020model,dwork2006calibrating}. Efforts to integrate DP within distributed learning environments, particularly in federated learning, have been increasing~\cite{fernandez2022privacy,nguyen2022federated}. Abadi et al.~\cite{abadi2016deep} developed a seminal approach called Deep Learning with Differential Privacy (DP-SGD), which adapts the Stochastic Gradient Descent (SGD) algorithm to meet DP standards by clipping gradients and injecting noise, thereby offering stringent privacy safeguards during deep neural network (DNN) training. Building on this, McMahan et al.~\cite{mcmahan2017learning} further tailored DP mechanisms for federated learning through an extension called DP-FedAvg. While these methods effectively address privacy concerns, they often fall short in dealing with dynamic data distributions, a prevalent issue in distributed learning~\cite{chen2021unified}. Specifically, data sets can evolve over time, rendering some information outdated or irrelevant, and the persistence of such data in the learning process can compromise model efficacy. Although Machine Unlearning approaches like Sharded, Isolated, Sliced, and Aggregated (SISA) training~\cite{bourtoule2021machine} have emerged to tackle this issue by enabling efficient selective forgetting of data, these methods are not yet designed to work synergistically with privacy-preserving techniques like DP~\cite{jegorova2022survey}.

Federated learning has substantially revolutionized distributed learning, enabling the training of Machine Learning models on decentralized networks while preserving data privacy and minimizing communication costs~\cite{zhou2019privacy}. Among the pioneering works in this area is the FedAvg algorithm by McMahan et al.~\cite{mcmahan2017communication}, which relies on model parameter averaging across local models and a central server. However, FedAvg is not without its limitations, particularly when handling non-IID data distributions~\cite{ma2023flgan}. Solutions like FedProx by Li et al.~\cite{li2020federated} have sought to address this by introducing a proximal term for improved model convergence. While other researchers like Sahu et al.~\cite{Sahu2018OnTC} and Konečný et al.~\cite{konevcny2016federated} have made strides in adaptive learning rates and communication efficiency, the realm of federated learning still faces significant challenges in dynamic adaptability and efficient Machine Unlearning. While privacy has been partially addressed through Differential Privacy~\cite{zhu2020more} and Secure Multiparty Computation~\cite{bonawitz2017practical}, these techniques often compromise on model efficiency. Additionally, the applicability of federated learning in diverse sectors like healthcare and IoT emphasizes the unmet need for a model capable of dynamically adapting to varied data distributions, while preserving privacy and efficiency~\cite{shaik2022fedstack,brisimi2018federated}.

Reinforcement Learning has garnered much attention for its ability to train agents to make optimal decisions through trial-and-error interactions with their environments~\cite{huang2021feddsr, hospedales2021meta}. Several pivotal advancements have shaped the field, including the development of Deep Q-Networks (DQNs)~\cite{mnih2013playing}. DQNs combine traditional reinforcement learning techniques with DNNs, significantly enhancing the system's ability to process high-dimensional inputs such as images. Furthermore, experience replay mechanisms have been integrated into them to improve learning stability by storing and reusing past experiences~\cite{schaul2015prioritized}. Mnih et al.~\cite{mnih2015human} significantly accelerated the reinforcement learning field by implementing DQNs that achieved human-level performance on a variety of complex tasks. However, there are evident gaps in addressing challenges posed by non-stationary or dynamic environment situations where the statistical properties of the environment change over time. Under such conditions, a reinforcement learning agent's ability to adapt quickly is paramount. Several approaches like meta-learning~\cite{bing2022meta} and attention mechanisms~\cite{brauwers2021general, sorokin2015deep} have sought to remedy these issues to some extent. Meta-learning, for example, helps models quickly adapt to new tasks by training them on a diverse range of tasks. However, the technique does not offer a robust solution for unlearning or forgetting outdated or irrelevant information, which is crucial for maintaining performance in dynamic environments. In a similar vein, attention mechanisms help agents focus on important regions of the input space, but they also fail to address the need for efficient unlearning of obsolete or irrelevant data. This leaves us with a significant research gap: the lack of mechanisms for efficient unlearning and adaptability in reinforcement learning agents designed for non-stationary, dynamic environments.

A key challenge for federated learning when faced with dynamic data distributions and the accumulation of outdated or irrelevant information is its adaptability in evolving environments. Reinforcement learning has been instrumental in training agents for optimal decision-making in dynamic environments, yet it too grapples with the need to efficiently unlearn outdated or irrelevant data. These challenges underscore the importance of integrating attention mechanisms into the Machine Unlearning process. Unlike selective data deletion, attention mechanisms assign reduced weights to outdated, private, or irrelevant information. The dynamic adjustment of attention scores allows these models to prioritize relevant data while disregarding obsolete or extraneous elements. By bridging the worlds of federated learning and reinforcement learning with attention mechanisms, our study addresses the pressing need for an integrated solution that optimizes decision-making in distributed networks with changing data landscapes~\cite{guan2023rfdg}. In addition, this approach must preserve data privacy and adaptively forget outdated, private, or irrelevant information.

\section{Preliminaries \& Problem Definition}\label{problem}  

This section establishes the foundational concepts and mathematical notations essential for the discussions and analyses presented in this paper. These concepts are summarized in Tab.~\ref{table:notations-merged} and form the basis for understanding the subsequent problem definitions and solution approaches. Our research is centered around the exploration of unlearning mechanisms in Machine Learning models, focusing on maintaining accuracy and computational efficiency while addressing the challenges posed by outdated or irrelevant data.

The problem is defined by two distinct settings: single-modality and multimodality. The single-modality setting is simpler and widely applicable in scenarios with uniform data types, such as sensor networks or content recommendation systems. However, it may lack the context provided by different types of data, potentially leading to less nuanced decisions. On the other hand, the multimodality setting is more complex but highly relevant in fields like healthcare, where a range of data types (e.g., medical imaging, patient history, etc.) can be used for more comprehensive understanding and decision-making. By exploring the problem in both these settings, we offer solutions that are both versatile and contextually rich.

\begin{table}[h]
\centering
\caption{Summary of Notations and Descriptions}
\label{table:notations-merged}
\resizebox{0.8\columnwidth}{!}{%
\begin{tabular}{|l|l|}
\hline
\textbf{Symbol} & \textbf{Description} \\ \hline
$AG$ & Set of agents in the model \\ \hline
$ag$ & An individual agent in the set $AG$ \\ \hline
$S_{i}$ & States observed by an agent $ag$ \\ \hline
$A$ & Set of possible actions \\ \hline
$\pi_{i}(s,a)$ & Policy followed by the agent \\ \hline
$R_{i}(s,a)$ & Rewards for actions in different states \\ \hline
$\theta_{ag}$ & Parameters of local models for agent $ag$ \\ \hline
$\theta_{g}$ & Parameters of global model \\ \hline
$w_{i,ag}$ & Attention score for a data point $i$ in agent $ag$ \\ \hline
$M$ & Set of modalities in multimodality setting \\ \hline
$X_{m}$ & Data vectors for modality $m$ \\ \hline
$\theta_{m}$ & Parameters for modality $m$ \\ \hline
$w_{i,m}$ & Attention scores within a modality $m$ \\ \hline
$t$ & Time step \\ \hline
$s_t$ & State at time step $t$ \\ \hline
$a_t$ & Action at time step $t$ \\ \hline
$r_t$ & Reward at time step $t$ \\ \hline
$R_t$ & Cumulative reward \\ \hline
$\pi(a_t|s_t)$ & Policy function \\ \hline
$Q(s_t,a_t)$ & $Q$-function \\ \hline
$\gamma$ & Discount factor \\ \hline
$\alpha_i$ & Attention score for feature $i$ \\ \hline
$\Delta\theta_{ag}$ & Update sent by agent $ag$ \\ \hline
$f$ & Function for calculating attention scores \\ \hline
$w_{g,ag}$ & Global attention score for update from agent $ag$ \\ \hline
$AG$ & Number of local agents \\ \hline
$\alpha_{\text{avg}}$ & Average attention score \\ \hline
$\delta$ & Predetermined threshold for attention score \\ \hline
$ag \in AG$ & A specific agent within the set of all agents $AG$ \\ \hline
$m$ & Number of modalities \\ \hline
$x_1, x_2,...,x_m$ & Data vectors for each modality \\ \hline
$v_i$ & Feature vector for modality $i$ \\ \hline
$\bar{w}_j$ & \begin{tabular}[c]{@{}l@{}}Averaged attention score across\\ modalities for data point $j$\end{tabular} \\ \hline
$\lambda$ & Mixing factor \\ \hline
$T$ & The total number of training rounds \\ \hline
$\alpha$ & Learning rate for $Q$-value function updates \\ \hline
$\eta$ & Scaling factor for attention score updates \\ \hline
$\beta$ & \begin{tabular}[c]{@{}l@{}}Mixing factor for combining \\ global and local model parameters\end{tabular} \\ \hline
$\varepsilon$ & Convergence threshold for global model parameters \\ \hline
$w_{ag}$ & Local model parameters for agent $ag$ \\ \hline
$W$ & Global model parameters \\ \hline
$A_i$ & Attention score for data point $i$ \\ \hline
$A_{i,ag}$ & Attention score for data point $i$ within agent $ag$ \\ \hline
$N$ & Total number of data points across all agents \\ \hline
$n_{ag}$ & Number of data points in agent $ag$ \\ \hline
\end{tabular}}
\end{table}


\begin{figure}
    \centering
    \includegraphics[scale =0.5]{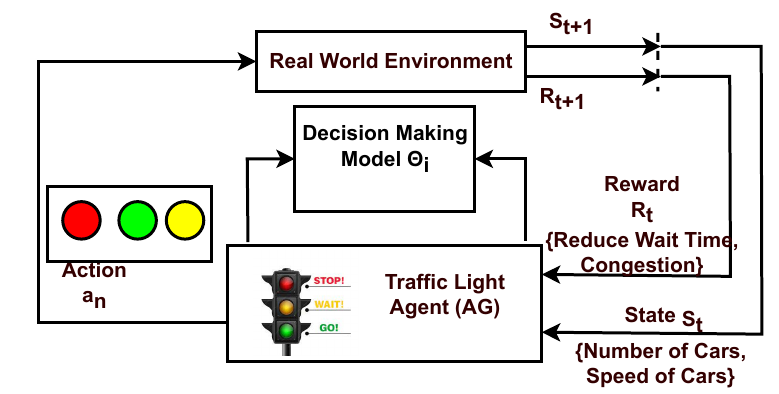}
    \caption{Single Modality Example}
    \label{fig:prob_1}
\end{figure}
\subsection{Problem Definition - Single Modality}

\begin{Problem Definition}
\textit{Let \( AG = \{ ag_1, ag_2, \ldots, ag_n \} \) be a set of agents, where each agent \( ag \in AG \) represents an entity like an IoT device, traffic point, wearable device, edge computing node, or content recommendation system. Each agent \( ag \) observes states \( S_i = \{ s_1, s_2, \ldots, s_m \} \) and takes actions \( A = \{ a_1, a_2, \ldots, a_n \} \) based on a policy \( \pi_i(s, a) \). Rewards \( R_i(s, a) \) evaluate the quality of actions taken in different states. Agents possess local models with parameters \( \theta_i \), while a central server maintains a global model with parameters \( \theta_g \).}
\end{Problem Definition}

\begin{example}
\textit{In the single-modality setting shown in Fig.~\ref{fig:prob_1}, let \( AG = \{ ag_1, ag_2, \ldots, ag_n \} \) be a set of agents. An agent \( ag \) can represent a real-world entity such as a traffic light in a city. These traffic lights observe various states \( S_i = \{ s_1, s_2, \ldots, s_m \} \), such as the number and speed of passing cars, and the change of colors (actions \( A = \{ a_1, a_2, \ldots, a_n \} \)) according to an algorithmic policy \( \pi_i(s, a) \). The system evaluates the effectiveness of the traffic light changes in reducing wait time or congestion (rewards \( R_i(s, a) \)). Each traffic light has its own local decision-making model characterized by parameters \( \theta_i \), and there is a global model for optimizing city-wide traffic flow with parameters \( \theta_g \).}
\end{example}

To address the challenge of preserving data privacy and adaptively forgetting private, outdated, or irrelevant information, attention scores $w_{ij}$ are assigned to each data point $j$ in the local dataset of agent $ag \in AG$. These attention scores, computed using a function $f$ that considers the current model state or contextual information, guide the learning and unlearning process within each agent. By assigning higher attention scores to relevant data and potentially forgetting or down-weighting irrelevant data, the agents can effectively focus on the most informative and up-to-date information.

\begin{figure}
    \centering
    \includegraphics[scale=0.5]{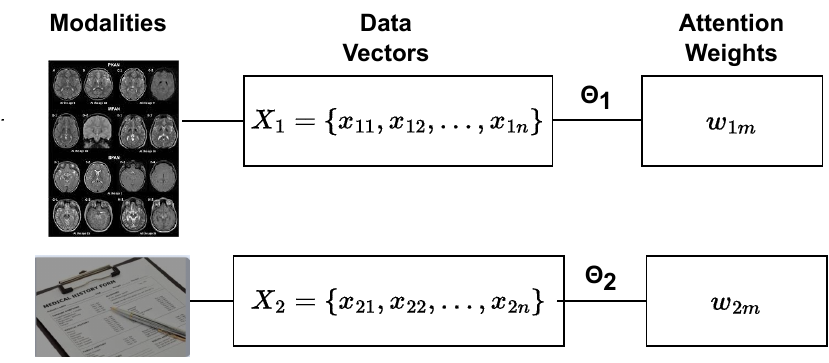}
    \caption{Multimodality Example}
    \label{fig:prob_2}
\end{figure}

\subsection{Problem Definition - Multimodality}

\begin{Problem Definition}
\textit{In the multimodality setting, let \( M = \{ 1, 2, \ldots, m \} \) represent the set of modalities, where \( m \) is the total number of modalities. Each modality \( m \in M \) is associated with a set of data vectors \( X_m = \{ x_{m1}, x_{m2}, \ldots, x_{mn} \} \), and has its local model with parameters \( \theta_k \). Attention scores \( w_{im} \) are assigned to individual data points \( x_{im} \) within each modality to guide the learning and unlearning process.}
\end{Problem Definition}

\begin{example}
\textit{In the multimodality setting shown in Fig.~\ref{fig:prob_2}, consider a healthcare system as a collection of agents in set \( M = \{ 1, 2, \ldots, m \} \), where \( m \) represents different types of medical data (modalities) such as medical imaging and patient history. For instance, medical imaging (modality \( M_1 \)) would have a set of MRI scans represented as data vectors \( X_1 = \{ x_{11}, x_{12}, \ldots, x_{1n} \} \). Likewise, patient history (modality \( M_2 \)) might involve a set of past diagnosis records that are represented as data vectors \( X_2 = \{ x_{21}, x_{22}, \ldots, x_{2n} \} \). Each modality has a specialized model with parameters \( \theta_1 \) for medical imaging and \( \theta_2 \) for patient history. These models use attention mechanisms to weigh the importance of each data point, represented by attention scores \( w_{1m} \) for MRI scans and \( w_{2m} \) for patient history records. These scores guide the decision-making process in diagnosis and treatment.}
\end{example}

In the multimodality setting, the complexity is elevated by the integration of heterogeneous data types and the application of specialized machine learning models for each modality. For example, in a healthcare system, combining data from disparate sources like medical imaging and patient history presents a unique challenge. Each data type, or modality, not only varies in format but also in the nature of the information it conveys, necessitating distinct processing and analysis methods. The key challenge here is to develop an integrated approach that effectively synthesizes these diverse data streams into a coherent understanding, enhancing decision-making in critical applications such as patient diagnosis and treatment. Attention mechanisms play a crucial role in this context, determining the relevance of each data point across different modalities. However, assigning and calibrating these attention scores is non-trivial and introduces an additional layer of complexity. The successful implementation of multimodal systems has profound implications, particularly in improving the accuracy and efficacy of decision-making processes. 


\begin{figure*}
\centering
\includegraphics[width=0.9\textwidth]{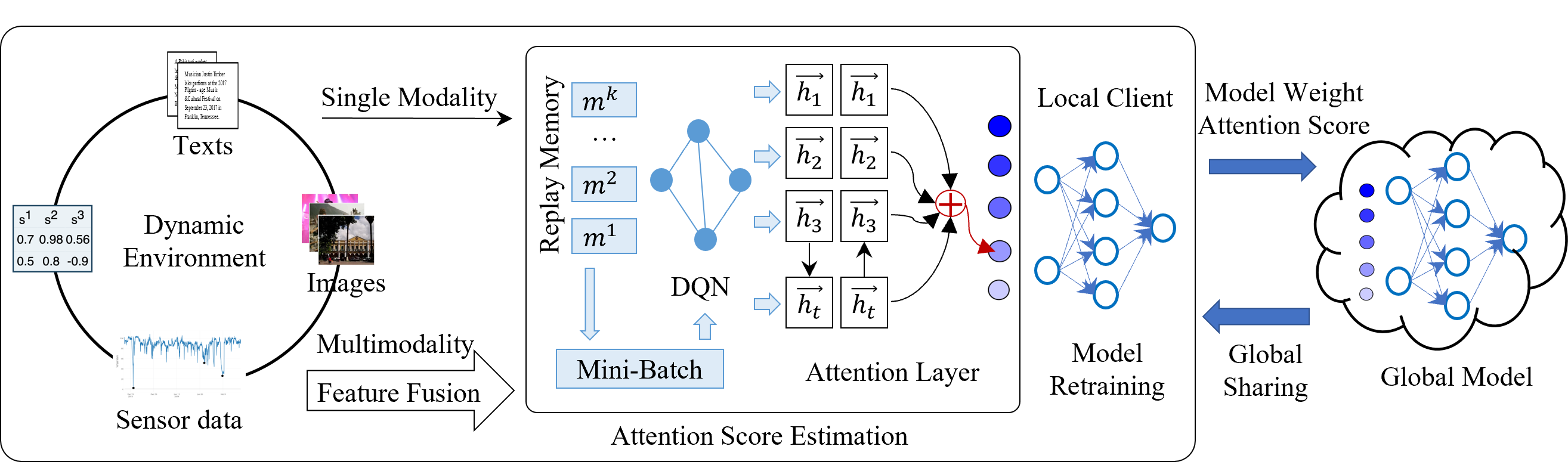}
\caption{An overview of the proposed FRAMU framework, illustrating its end-to-end adaptive algorithm that incorporates an attention mechanism. The figure is divided into multiple components, each corresponding to a specific phase in the federated learning process. Starting from the left, the diagram begins with data collection from diverse modalities. The framework applies an adaptive learning algorithm that not only updates the global model, but also incorporates an efficient unlearning mechanism for discarding outdated, private, or irrelevant data.}
\label{fig:FRAMU_Framework}
\end{figure*}

\section{FRAMU Framework}\label{methods}

In an era marked by an ever-increasing influx of data, the need for adaptive Machine Learning models that can efficiently unlearn outdated, private, or irrelevant information is paramount. The methodology proposed in this paper addresses this necessity by introducing two key technical contributions. First, we propose an adaptive unlearning algorithm that utilizes attention mechanisms to tailor the learning and unlearning processes in a single-modality, and then extend the process to multimodality. This innovative approach allows the model to adapt to dynamic changes in data distributions, as well as variations in participant characteristics such as demographic information, behavioural patterns, and data contribution frequencies among others. Second, we put forth a novel design that employs the FedAvg mechanism~\cite{mcmahan2017communication} to personalize the unlearning process. This design ensures that the model is able to discard data that has become irrelevant, outdated, or potentially invasive from a privacy perspective, thus preserving the integrity of the learning model while adapting to new or changing data. The following sections will elaborate on these contributions, providing a detailed discussion of the proposed framework as depicted in Fig.~\ref{fig:FRAMU_Framework}.

The FRAMU framework adopts a federated learning architecture comprising Local Agents and a Central Server, each with distinct roles in model training, unlearning, and adaptation. It employs a reinforcement learning paradigm where each agent iteratively learns from its environment. This integration of federated learning and reinforcement learning is termed federated reinforcement learning. However, what sets FRAMU apart is the integration of attention mechanisms to weigh the relevance of each data point in learning and unlearning. The attention scores are then aggregated and processed at the Central Server to refine the global model.


\begin{itemize}
    \item \textbf{Local Agents}: Responsible for collecting real-time data and performing local model updates. They observe states, take actions, and calculate rewards to update their Q-values and attention scores.
    \item \textbf{Central Server}: Aggregates local models and attention scores, filters out irrelevant data points, and updates the global model.
    \item \textbf{Attention Mechanism}: Dynamically calculates attention scores for each data point to inform the unlearning process.
    \item \textbf{FedAvg Mechanism}: Utilized for global model updates, ensuring that the global model represents a consensus across all agents.
\end{itemize}

\begin{algorithm}[t!]
\small
\caption{\textbf{FRAMU Framework}} 
\label{alg:framu}
\KwIn{a set of Local Agents, a Central Server, $T$, $\theta$, $\alpha$, $\eta$, $\gamma$, $\beta$, $\varepsilon$}
\KwOut{ $\widehat{W}:$ Trained global model parameters for federated reinforcement learning}
\BlankLine
Initialize local model parameters $w_{ag}$ for each agent $ag$; \\
Initialize global model parameters $W$ at the central server; \\
Initialize attention scores $A_{i,ag,m}$ for each data point $i$ in agent $ag$ and modality $m$; \\
\While{$t \leq T$}{
\ForEach{local agent $ag$}{
    Observe current states $s_{i,m}$ for each modality $m$; \\
    Take action $a_t$ based on policy derived from $Q(s, a; w_{ag})$; \\
    Observe reward $r_t$ and next states $s_{i,m}'$ for each modality $m$; \\
    Compute TD error $\delta = r_t + \gamma \max_{a} Q(s_{i,m}', a; w_{ag}) - Q(s_{i,m}, a_t; w_{ag})$; \\
    Update $Q(s_{i,m}, a_t; w_{ag}) \leftarrow Q(s_{i,m}, a_t; w_{ag}) + \alpha \delta$; \\
    Update attention scores $A_{i,ag,m} \leftarrow A_{i,ag,m} + \eta |\delta|$; \\
}
Send local model parameters $w_{ag}$ and attention scores $A_{i,ag,m}$ to Central Server; \\
\ForEach{data point $i$ in modality $m$}{
    \If{$\sum_{ag} \frac{1}{m}\sum_{m} A_{i,ag,m} / N_{ag} < \theta$}{
        Reduce influence of data point $i$ in the global model; \\
    }
}
Aggregate local model parameters to update global parameters: $W \leftarrow \sum_{ag} \left(\frac{n_{ag}}{N}\right) w_{ag}$; \\
Send updated global model parameters $W$ to local agents; \\
\ForEach{local agent $ag$}{
    Fine-tune local model with global model: \\
    $w_{ag}' \leftarrow \beta W + (1 - \beta) w_{ag}$; \\
}
\If{$|P(W_{t+1}) - P(W_t)| < \varepsilon$}{
    Break; \\
}
Increment $t$; \\
}
\Return{$W$} \label{alg1:result}
\end{algorithm}

The FRAMU framework, as outlined in Algorithm~\ref{alg:framu}, has been carefully designed to facilitate adaptive decision-making in distributed networks through federated reinforcement learning. Each step within the algorithm is crafted with specific intentions: The initialization stage (Lines 1-3) sets the groundwork by initializing local and global model parameters, as well as attention scores. These initializations are crucial for ensuring that both local and global perspectives are considered right from the start of the learning process. The iterative learning process (Lines 4-24) involves several key components. Local Agent Decision-Making (Lines 5-11) enables each local agent to observe states, take actions, and update its Q-values and attention scores, ensuring that local knowledge is continuously updated to reflect the dynamic nature of the agents' environments. Central Server Aggregation (Lines 12-17) plays a pivotal role in integrating local updates and refining the global model. By assessing the attention scores, the server can identify and diminish the influence of less relevant data points, thereby enhancing the model's focus on significant information. Model Synchronization (Lines 18-24) involves the dissemination of global model parameters back to local agents for fine-tuning, ensuring a bi-directional flow of information that keeps local models informed by their immediate environment and aligned with the broader objectives of the global model. 



\section{Applications of FRAMU}\label{applicationsofframu}

This section explores the practical applications of the FRAMU framework across different settings, single-modality and multimodality, and its continuous adaptation and learning.

\subsection{FRAMU with Single Modality}
Central to FRAMU is an attention layer that functions as a specialized approximator, augmenting the learning capability of individual agents. This attention layer distinguishes itself by assigning attention scores to individual data points during the function approximation process. These scores serve as indicators of each data point's relevance in the agent's local learning. The agent updates these scores as it interacts with its environment and receives either rewards or penalties, thereby continually refining its model. Specifically, an agent operates in discrete time steps, current state $s_{t}$, taking action $a_{t}$, and receiving reward $r_{t}$, at each time step $t$. The ultimate goal is to determine an optimal policy $\pi(a_{t}|s_{t})$ that maximizes the accumulated reward $R_{t}$. The \textit{Q}-function, which quantifies expected accumulated rewards with a discount factor $\gamma$, is given by Equation~\ref{Q_function}.

\begin{equation}\label{Q_function}
Q(s_{t}, a_{t}) = \mathbb{E}[R_{t} \mid s_{t}, a_{t}] = r_{t} + \gamma \mathbb{E}[Q(s_{t+1}, a_{t+1}) \mid s_{t}, a_{t}]
\end{equation}

The attention layer further characterizes each state $s_t$ by its features $[x_1, x_2, ..., x_n]$, and assigns attention scores $\alpha_i$ as per:

\begin{equation}
\alpha_i = \text{Attention}(x_i, \text{context})
\end{equation}

Here, the context may include additional data such as previous states or actions. The \textit{Q}-function is then approximated using a weighted sum of these features:

\begin{equation}
Q(s_{t}, a_{t}) \approx \sum (\alpha_{i} \cdot x_{i})
\end{equation}

After completing their respective learning cycles, agents forward their model updates $\theta$ and attention scores $\alpha$ to the Central Server as a tuple $(\theta, \alpha)$.

\subsubsection{Local and Global Attention Score Estimation}
FRAMU estimates attention scores both locally and globally. On the local front, each agent employs its attention mechanism to compute scores for individual data points based on their relevance to the task at hand. For an agent $ag$ with local model parameters $\theta_{ag}$, the attention score $w_{ij}$ for data point $j$ is given by:

\begin{equation}
w_{ij} = f(s_{j}, \theta_{ag})
\end{equation}

At the global level, these scores assist the Central Server in prioritizing updates or pinpointing data points for global unlearning. For global parameters $\theta_g$, the global attention score derived from the updates of agent $ag$ is:

\begin{equation}
w_{g,ag} = f(\Delta\theta_{ag}, \theta_{g})
\end{equation}

In this equation, $\Delta\theta_{ag}$ is the model update from agent $ag$, and the function $f$ calculates attention scores while taking into account the aggregated local scores and other global contextual cues.

\subsubsection{Global Model Refinement and Unlearning}
Model updates from local agents are aggregated at the Central Server using FedAvg~\cite{miao2023task}. The attention scores are instrumental in the global unlearning process, with the average attention score calculated as:

\begin{equation}
\alpha_{\text{avg}} = \frac{1}{AG} \sum \alpha_{ag}
\end{equation}

When $\alpha_{\text{avg}}$ falls below a predetermined threshold $\delta$, the server adjusts the contribution of the respective feature in the global model as given by Equation~\ref{global_model}:

\begin{equation}\label{global_model}
\theta_{\text{global'}} = g(\theta_{\text{global}}, \alpha_{\text{avg}})
\end{equation}

Once refined, this global model is sent back to the local agents. The enhanced model shows improved adaptability and robustness to changes in data distributions due to the integration of aggregation and unlearning mechanisms. Consequently, the local agents are better positioned to excel within their particular operational environments. These revised global model parameters, denoted as $\theta_{\text{global'}}$, are then dispatched from the Central Server to the local agents, where $\theta_k = \theta_{\text{global'}}$.

\subsection{FRAMU with Multimodality}

The multimodal FRAMU Framework extends its capabilities to seamlessly incorporate various data types, including images, text, audio, and sensor readings. This integration not only enriches decision-making but also optimizes the performance of local agents. By fine-tuning their models to multiple data types, agents are better equipped to operate in complex environments.

\subsubsection{Modality-Specific Attention Mechanisms}

To effectively manage data from diverse sources, the framework employs specialized attention mechanisms for each modality. These mechanisms generate unique attention scores for data points within a given modality, aiding in both learning and unlearning processes. By doing so, the framework allows local agents to focus on the most relevant and informative aspects of each modality.

The attention scores for a specific modality \( j \) for an agent \( \text{ag} \in \text{AG} \) can be mathematically represented as:

\begin{equation}
w_{ij} = f_j(s_{ij}, \theta_i),
\end{equation}

Here, \( s_{ij} \) signifies a data point from modality \( j \) related to agent \( \text{ag} \in \text{AG} \), while \( \theta_i \) represents that agent's local model parameters. The function \( f_j \) considers modality-specific attributes and context to compute these attention scores.

For a feature vector \( v_i \) derived from modality \( j \) within agent \( \text{ag} \in \text{AG} \), feature-level fusion can be represented as:

\begin{equation}\label{feature_vector}
v_i = [x_{i1}, x_{i2}, \ldots, x_{im}]
\end{equation}

\subsubsection{Unlearning and Adaptation across Modalities}

In a multimodal setup, attention scores from all modalities collectively inform the unlearning process. If a data point consistently receives low attention scores across different modalities, it indicates that the point is either irrelevant or outdated. The Central Server uses this multimodal insight to refine the global model.

The average attention score across all modalities for a specific data point is:

\begin{equation}
\bar{w}_j = \frac{1}{m}\sum_{i=1}^{m} w_{ij}
\end{equation}

If \( \bar{w}_j \) falls below a predefined threshold, the Central Server de-emphasizes or removes that data point from the global model, ensuring that only current and relevant data contribute to decision-making.

During the adaptation phase, local agents utilize the updated global model to enhance their local models. The interplay between global and local parameters is regulated by a mixing factor, which allows local agents to leverage shared insights while preserving modality-specific skills. This relationship can be denoted by:

\begin{equation}
\theta_{i}^{\text{new}} = \lambda \theta_{\text{global}} + (1-\lambda) \theta_{i}^{\text{old}}
\end{equation}

Here, \( \theta_{i}^{\text{new}} \) represents the updated local model parameters, \( \theta_{\text{global}} \) signifies the global model parameters, \( \theta_{i}^{\text{old}} \) is the previous local parameters, and \( \lambda \) serves as the mixing factor. Through this, the multimodal FRAMU framework maintains an up-to-date and relevant global model, while enabling local agents to make better decisions across a range of data types.

\subsection{Continuous Adaptation and Learning in the FRAMU Framework}

Continuous adaptation and learning are critical in the FRAMU framework, enabling it to thrive in dynamic and changing environments. These processes create an iterative exchange of knowledge between local agents and a Central Server, which leads to consistent model refinement on both local and global scales.

\subsubsection{Local-Level Adaptation}

Local agents need the ability to adapt in real time to changes in their operational environments. Within reinforcement learning paradigms, agents continually update their policies in response to actions taken and rewards observed. Furthermore, attention scores allocated to data points or features can vary dynamically based on new data or shifts in relevance. This adaptability ensures that the models of individual local agents remain current. Let \( s_t \) denote the state of the environment at time \( t \), and \( a_t \) represent the action taken by the agent. After receiving a reward \( r_t \) and transitioning to a new state \( s_{t+1} \), the agent aims to maximize the expected cumulative reward. The Q-value function \( Q(s, a) \) serves as a proxy for this cumulative reward, and it is updated using temporal-difference learning algorithms as follows:

\begin{equation}
Q(s_t, a_t) \leftarrow Q(s_t, a_t) + \alpha \left[ r_t + \gamma \max_a Q(s_{t+1}, a) - Q(s_t, a_t) \right]
\end{equation}

Here, \( \alpha \) is the learning rate, and \( \gamma \) is the discount factor.

Attention scores, denoted by \( A_i \) for data point \( i \), are updated based on the temporal-difference error \( \delta \):

\begin{equation}
A_i \leftarrow A_i + \eta |\delta|,
\end{equation}

where \( \eta \) is a scaling factor, and \( \delta = r_t + \gamma \max_a Q(s_{t+1}, a) - Q(s_t, a_t) \).

\subsubsection{Global Model Aggregation and Adaptation}

As local agents continuously update their models, these adaptations are communicated to the Central Server. It aggregates this information to refine the global model while also tracking the attention scores from local agents. If these scores reveal diminishing importance for certain data points, the server may initiate global unlearning. This ensures the global model remains current and avoids obsolescence. Local agents send their updated model parameters, \( w_{ag} \) for agent \( ag \), and attention scores \( A_{i,ag} \) to the Central Server. The server aggregates these to update the global model parameters \( W \) as follows:

\begin{equation}
W \leftarrow \frac{1}{AG} \sum_{ag} w_{ag},
\end{equation}

where \( AG \) represents the total number of local agents.

\subsubsection{Feedback Mechanisms}

After the global model is updated, it is disseminated back to local agents through a feedback loop. This cyclic interaction allows local agents to either initialize or further refine their models based on the global one. This is particularly beneficial when local agents confront new or unfamiliar data points that other agents have encountered. Through this mechanism, the global model acts as a repository of shared knowledge, enhancing the decision-making capabilities of all local agents. The global model parameters \( W \) are sent to local agents, who then adjust their local models using a mixing factor \( \beta \) as follows:

\begin{equation}
w_{ag}' \leftarrow \beta W + (1 - \beta) w_{ag},
\end{equation}

where \( \beta \) ranges from 0 to 1 and regulates the influence of the global model on local models.

\begin{figure}
    \centering
    \includegraphics[scale=0.4]{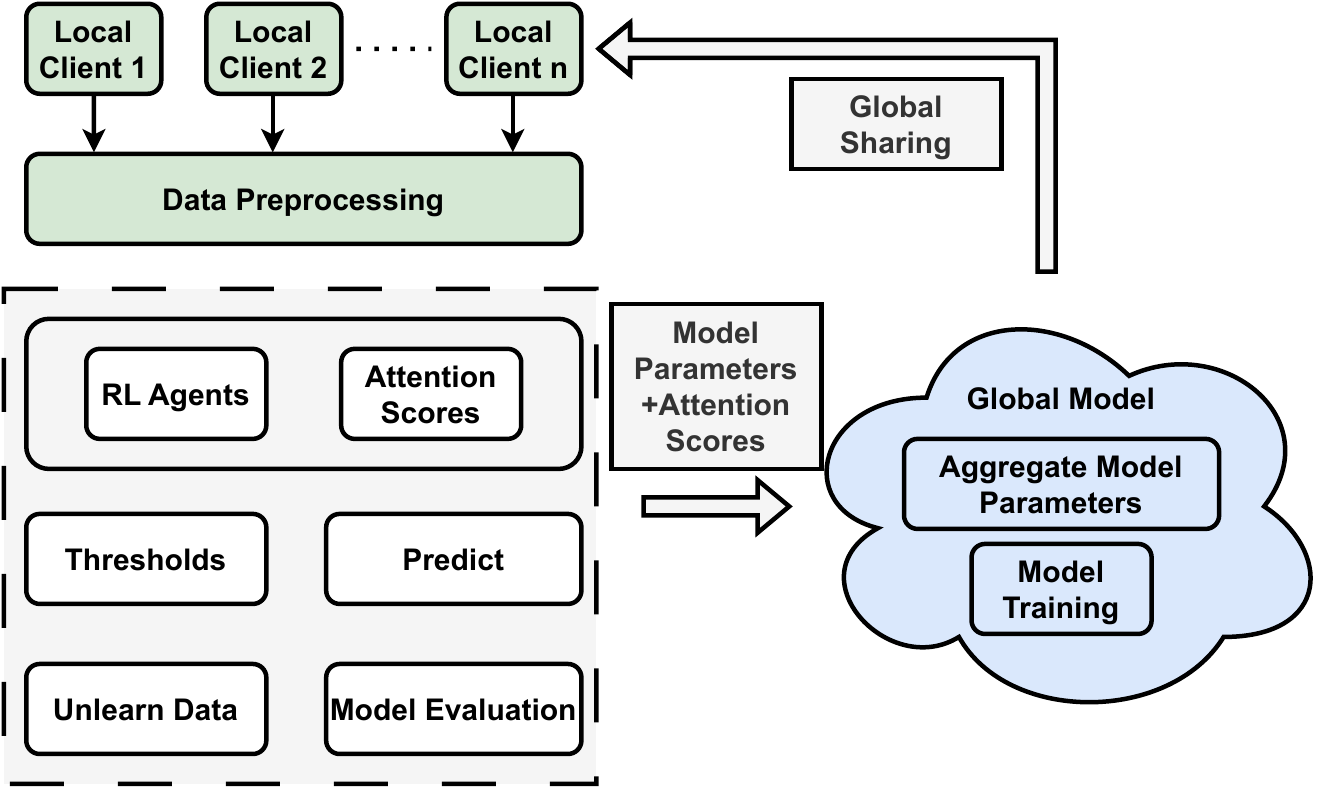}
    \caption{Experimental Setup: This diagram showcases the architecture of the FRAMU framework, detailing the interaction between local and global models within a federated learning environment.}
    \label{fig:experiemental_design}
\end{figure}

\section{Experimental Setup and Results Analysis}\label{experiment}

To effectively evaluate the performance of the FRAMU framework, we undertook comprehensive experiments using real-world datasets. These experiments were designed to validate not only the efficiency and effectiveness of our approach but also to establish the practical utility of FRAMU in real-world applications. Our experimental setup encompassed several components, including datasets, baseline models, evaluation metrics, and specific FRAMU configurations, as depicted in Fig.~\ref{fig:experiemental_design}. A critical aspect of our experimentation involved fine-tuning key thresholds to guide the unlearning process, particularly the outdated\_threshold and irrelevant\_threshold. These parameters were adjusted based on domain expertise and sensitivity analysis, with the outdated\_threshold defining the time frame for data obsolescence and the irrelevant\_threshold setting criteria for data's statistical insignificance. Additionally, we introduced a privacy\_epsilon parameter to balance data utility with privacy preservation, aligning with GDPR regulations.

Deep learning methods are known for their ability to learn features autonomously and automate model-building processes. Despite criticisms of neural network family algorithms for their 'black box' nature, deep learning models are renowned for their robust and efficient performance. These models are widely adopted by the research community~\cite{ek2021federated, ouyang2021clusterfl}. In our work, we utilized a Convolutional Neural Network (CNN) for image and sensor data, and a Long Short-Term Memory (LSTM) network for time series and text data, specifically tailored for federated learning scenarios. This model choice was made to efficiently handle both single and multimodal data types, integrating attention mechanisms and unlearning processes to enhance overall functionality.

We found the tuning of parameters such as the outdated threshold, irrelevant threshold, and the $\beta$ value for local model fine-tuning to be crucial. It was essential to strike the right balance in the outdated threshold to prevent premature data discarding or retention of outdated information, which could affect model accuracy and relevancy. Similarly, careful calibration of the irrelevant threshold was necessary to maintain a balance between data comprehensiveness and quality, ensuring useful data was not excluded nor excessive noise retained. The $\beta$ value, crucial in determining the extent of global model influence on local models, required fine-tuning to ensure an optimal balance between local and global learning. This was key for local models to benefit from global insights while preserving their unique learning characteristics. The interplay of these hyperparameters significantly influenced FRAMU's performance, particularly in its ability to adapt to new data and retain relevant historical information. Through sensitivity analyses, we determined their optimal ranges, aiming to maximize FRAMU's efficiency and adaptability in various real-world scenarios. 

\begin{table}[hbt!]
\caption{Datasets for evaluation}
\label{tab:datasets}
\resizebox{\columnwidth}{!}{
\begin{tabular}{|c|c|c|c|c|c|}
\hline
\textbf{Modality} & \textbf{Dataset} & \textbf{\begin{tabular}[c]{@{}c@{}}OD*\end{tabular}} & \textbf{\begin{tabular}[c]{@{}c@{}}PD*\end{tabular}} & \textbf{\begin{tabular}[c]{@{}c@{}}ID*\end{tabular}} & \textbf{Description} \\ \hline
\multirow{3}{*}{\textbf{\begin{tabular}[c]{@{}c@{}}Single\\ Modality\end{tabular}}} & \begin{tabular}[c]{@{}c@{}}AMPds2\\ \cite{makonin2016electricity}\end{tabular} & \cmark & \cmark & \cmark & \begin{tabular}[c]{@{}c@{}}Electricity, water, and\\ natural gas consumption data \\ .from a Canadian household.\end{tabular} \\ \cline{2-6} 
 & \begin{tabular}[c]{@{}c@{}}METR-LA\\ \cite{li2018dcrnn_traffic}\end{tabular} & \cmark & \xmark & \cmark & \begin{tabular}[c]{@{}c@{}}Traffic speed data from\\ over 200 sensors in Los\\ Angeles Metropolitan area.\end{tabular} \\ \cline{2-6} 
 & \begin{tabular}[c]{@{}c@{}}MIMIC-III\\ \cite{johnson2016mimic}\end{tabular} & \cmark & \cmark & \cmark & \begin{tabular}[c]{@{}c@{}}Health-related data from\\ critical care units,including\\ demographics, vital\\ signs, laboratory results,\\ and medications.\end{tabular} \\ \hline
\multirow{3}{*}{\textbf{\begin{tabular}[c]{@{}c@{}}Multi\\ Modality\end{tabular}}} & \begin{tabular}[c]{@{}c@{}}NYPD\\ \cite{(NYPD)_2023}\end{tabular} & \cmark & \cmark & \cmark & \begin{tabular}[c]{@{}c@{}}Records of complaints filed \\ with the New York City \\ Police Department.\end{tabular} \\ \cline{2-6} 
 & \begin{tabular}[c]{@{}c@{}}MIMIC-CXR\\ \cite{johnson2019mimic}\end{tabular} & \cmark & \cmark & \cmark & \begin{tabular}[c]{@{}c@{}}Chest radiographs with \\ associated radiology\\ reports for medical\\ image analysis tasks.\end{tabular} \\ \cline{2-6} 
 & \begin{tabular}[c]{@{}c@{}}Smart Home \\ EnergyDataset \\ (SHED)\\ \cite{Dataset_2019}\end{tabular} & \cmark & \cmark & \cmark & \begin{tabular}[c]{@{}c@{}}Energy consumption data\\ from smart home devices\\ and appliances.\end{tabular} \\ \hline
\end{tabular}}
\footnotesize\textsuperscript{*}OD - Outdated Data, PD - Privacy Data, ID - Irrelevant Data.
\end{table}

\subsection{Datasets}
In this study, publicly available datasets that encompass various modalities and address specific challenges related to outdated, private, and irrelevant data are adopted. Tab.~\ref{tab:datasets} provides detailed information about each dataset, including the data modality, number of instances, attributes, target variables, and specific characteristics pertinent to our study. In order to evaluate FRAMU, we conducted a comprehensive comparison of its performance against several contemporary baseline models.

\subsection{Baseline Models}

In the evaluation of the FRAMU framework's performance and robustness, we have carefully selected several baseline models for comparison. The models in baseline models were adopted from the original work. The rationale behind choosing each model and its relevance to our study is elaborated below:

\begin{itemize}
    \item \textbf{Single-modality}

\begin{itemize}
    \item \textbf{FedLU~\cite{zhu2023heterogeneous}}: FedLU represents a significant advance in federated learning, integrating knowledge graph embedding with mutual knowledge distillation. Its selection as a baseline is due to its innovative approach to collaborative learning, which is closely aligned with FRAMU’s objectives in single-modality settings. FedLU's methodology provides a comparative framework for assessing FRAMU's efficiency in knowledge synthesis and distribution.   
    \item \textbf{Zero-shot MU~\cite{chundawat2023zero}}: Zero-shot MU specializes in Machine Unlearning, employing error-minimizing-maximizing noise and gated knowledge transfer. This model was chosen for its novel approach to unlearning, providing a benchmark to evaluate FRAMU's capability in effectively removing learned information without extensive retraining, a crucial aspect in dynamic environments.    
    \item \textbf{SISA Training~\cite{bourtoule2021machine}}: The SISA Training framework is a strategic model that limits data points for optimized unlearning. Its inclusion as a baseline allows us to compare FRAMU's efficiency in data management and unlearning processes, especially in scenarios where data minimization is key to performance and privacy.
\end{itemize}

    \item \textbf{Multimodality}

\begin{itemize}
    \item \textbf{MMoE~\cite{ma2018modeling}}: The MMoE model, optimized for handling multimodal data via ensemble learning, serves as a benchmark for evaluating FRAMU's performance in multimodality settings. Its approach, employing expert networks for different data modalities, provides a comparative perspective for FRAMU's adaptability and efficiency in handling diverse data types.    
    \item \textbf{CleanCLIP~\cite{bansal2023cleanclip}}: CleanCLIP, a fine-tuning framework that mitigates spurious associations from backdoor attacks, is pivotal for comparing FRAMU's robustness against data security threats. Its focus on weakening spurious correlations offers insights into FRAMU's capabilities in maintaining data integrity and security.
    \item \textbf{Privacy-Enhanced Emotion Recognition (PEER)~\cite{jaiswal2020privacy}}: The PEER model, utilizing adversarial learning for privacy-preserving emotion recognition, aligns well with FRAMU’s privacy objectives. Its comparison with FRAMU highlights the effectiveness of FRAMU in safeguarding privacy while performing complex analytical tasks.
\end{itemize}

\end{itemize}

\begin{table*}[ht]
\scriptsize
\centering
\caption{FRAMU - Evaluation Results in Single Modality Context}
\label{tab:single_modality_mu}
\resizebox{\textwidth}{!}{%
\begin{tabular}{|c|cc|ccc|ccc|ccc|cc|}
\hline
\multirow{2}{*}{\textbf{Unlearning}} & \multicolumn{2}{c|}{\multirow{2}{*}{\textbf{Dataset}}} & \multicolumn{3}{c|}{\textbf{FedLU~\cite{zhu2023heterogeneous}}} & \multicolumn{3}{c|}{\textbf{Zero-shot~\cite{chundawat2023zero}}} & \multicolumn{3}{c|}{\textbf{SISA~\cite{bourtoule2021machine}}} & \multicolumn{2}{c|}{\textbf{FRAMU (Ours)}} \\ \cline{4-14} 
 & \multicolumn{2}{c|}{} & \multicolumn{1}{c|}{\textbf{MSE}} & \multicolumn{1}{c|}{\textbf{MAE}} & \textbf{p-value} & \multicolumn{1}{c|}{\textbf{MSE}} & \multicolumn{1}{c|}{\textbf{MAE}} & \textbf{p-value} & \multicolumn{1}{c|}{\textbf{MSE}} & \multicolumn{1}{c|}{\textbf{MAE}} & \textbf{p-value} & \multicolumn{1}{c|}{\textbf{MSE}} & \textbf{MAE} \\ \hline
\multirow{6}{*}{\textbf{\begin{tabular}[c]{@{}c@{}}Outdated\\ Data\end{tabular}}} & \multicolumn{1}{c|}{\multirow{3}{*}{\textbf{Original}}} & AMPds2 & \multicolumn{1}{c|}{0.063} & \multicolumn{1}{c|}{6.740} & 0.024 & \multicolumn{1}{c|}{0.061} & \multicolumn{1}{c|}{6.890} & 0.031 & \multicolumn{1}{c|}{0.059} & \multicolumn{1}{c|}{6.760} & 0.041 & \multicolumn{1}{c|}{\textbf{0.046}} & \textbf{5.570} \\ \cline{3-14} 
 & \multicolumn{1}{c|}{} & METR-LA & \multicolumn{1}{c|}{0.079} & \multicolumn{1}{c|}{7.140} & 0.016 & \multicolumn{1}{c|}{0.082} & \multicolumn{1}{c|}{7.210} & 0.038 & \multicolumn{1}{c|}{0.078} & \multicolumn{1}{c|}{7.090} & 0.029 & \multicolumn{1}{c|}{\textbf{0.065}} & \textbf{5.930} \\ \cline{3-14} 
 & \multicolumn{1}{c|}{} & MIMIC-III & \multicolumn{1}{c|}{0.099} & \multicolumn{1}{c|}{12.800} & 0.031 & \multicolumn{1}{c|}{0.102} & \multicolumn{1}{c|}{12.930} & 0.045 & \multicolumn{1}{c|}{0.097} & \multicolumn{1}{c|}{12.680} & 0.032 & \multicolumn{1}{c|}{\textbf{0.083}} & \textbf{10.650} \\ \cline{2-14} 
 & \multicolumn{1}{c|}{\multirow{3}{*}{\textbf{Unlearned}}} & AMPds2 & \multicolumn{1}{c|}{0.060} & \multicolumn{1}{c|}{6.630} & 0.015 & \multicolumn{1}{c|}{0.055} & \multicolumn{1}{c|}{6.860} & 0.029 & \multicolumn{1}{c|}{0.056} & \multicolumn{1}{c|}{6.690} & 0.036 & \multicolumn{1}{c|}{\textbf{0.038}} & \textbf{4.670} \\ \cline{3-14} 
 & \multicolumn{1}{c|}{} & METR-LA & \multicolumn{1}{c|}{0.075} & \multicolumn{1}{c|}{7.020} & 0.029 & \multicolumn{1}{c|}{0.077} & \multicolumn{1}{c|}{7.100} & 0.025 & \multicolumn{1}{c|}{0.072} & \multicolumn{1}{c|}{6.960} & 0.032 & \multicolumn{1}{c|}{\textbf{0.052}} & \textbf{4.910} \\ \cline{3-14} 
 & \multicolumn{1}{c|}{} & MIMIC-III & \multicolumn{1}{c|}{0.095} & \multicolumn{1}{c|}{12.650} & 0.023 & \multicolumn{1}{c|}{0.098} & \multicolumn{1}{c|}{12.820} & 0.041 & \multicolumn{1}{c|}{0.094} & \multicolumn{1}{c|}{12.580} & 0.017 & \multicolumn{1}{c|}{\textbf{0.069}} & \textbf{8.900} \\ \hline
\multirow{4}{*}{\textbf{\begin{tabular}[c]{@{}c@{}}Private\\ Data\end{tabular}}} & \multicolumn{1}{c|}{\multirow{2}{*}{\textbf{Original}}} & AMPds2 & \multicolumn{1}{c|}{0.052} & \multicolumn{1}{c|}{6.780} & 0.014 & \multicolumn{1}{c|}{0.054} & \multicolumn{1}{c|}{6.930} & 0.037 & \multicolumn{1}{c|}{0.053} & \multicolumn{1}{c|}{6.810} & 0.041 & \multicolumn{1}{c|}{\textbf{0.041}} & \textbf{5.540} \\ \cline{3-14} 
 & \multicolumn{1}{c|}{} & MIMIC-III & \multicolumn{1}{c|}{0.078} & \multicolumn{1}{c|}{12.870} & 0.035 & \multicolumn{1}{c|}{0.080} & \multicolumn{1}{c|}{13.010} & 0.043 & \multicolumn{1}{c|}{0.079} & \multicolumn{1}{c|}{12.760} & 0.045 & \multicolumn{1}{c|}{\textbf{0.064}} & \textbf{10.600} \\ \cline{2-14} 
 & \multicolumn{1}{c|}{\multirow{2}{*}{\textbf{Unlearned}}} & AMPds2 & \multicolumn{1}{c|}{0.049} & \multicolumn{1}{c|}{6.670} & 0.011 & \multicolumn{1}{c|}{0.052} & \multicolumn{1}{c|}{6.910} & 0.035 & \multicolumn{1}{c|}{0.051} & \multicolumn{1}{c|}{6.740} & 0.015 & \multicolumn{1}{c|}{\textbf{0.033}} & \textbf{4.590} \\ \cline{3-14} 
 & \multicolumn{1}{c|}{} & MIMIC-III & \multicolumn{1}{c|}{0.075} & \multicolumn{1}{c|}{12.720} & 0.031 & \multicolumn{1}{c|}{0.077} & \multicolumn{1}{c|}{12.900} & 0.038 & \multicolumn{1}{c|}{0.076} & \multicolumn{1}{c|}{12.650} & 0.016 & \multicolumn{1}{c|}{\textbf{0.053}} & \textbf{8.860} \\ \hline
\multirow{6}{*}{\textbf{\begin{tabular}[c]{@{}c@{}}Irrelevant\\ Data\end{tabular}}} & \multicolumn{1}{c|}{\multirow{3}{*}{\textbf{Original}}} & AMPds2 & \multicolumn{1}{c|}{0.047} & \multicolumn{1}{c|}{6.700} & 0.035 & \multicolumn{1}{c|}{0.050} & \multicolumn{1}{c|}{6.850} & 0.044 & \multicolumn{1}{c|}{0.048} & \multicolumn{1}{c|}{6.730} & 0.031 & \multicolumn{1}{c|}{\textbf{0.037}} & \textbf{5.440} \\ \cline{3-14} 
 & \multicolumn{1}{c|}{} & METR-LA & \multicolumn{1}{c|}{0.054} & \multicolumn{1}{c|}{7.100} & 0.027 & \multicolumn{1}{c|}{0.056} & \multicolumn{1}{c|}{7.170} & 0.041 & \multicolumn{1}{c|}{0.055} & \multicolumn{1}{c|}{7.050} & 0.025 & \multicolumn{1}{c|}{\textbf{0.043}} & \textbf{5.830} \\ \cline{3-14} 
 & \multicolumn{1}{c|}{} & MIMIC-III & \multicolumn{1}{c|}{0.070} & \multicolumn{1}{c|}{12.730} & 0.038 & \multicolumn{1}{c|}{0.072} & \multicolumn{1}{c|}{12.870} & 0.031 & \multicolumn{1}{c|}{0.071} & \multicolumn{1}{c|}{12.620} & 0.039 & \multicolumn{1}{c|}{\textbf{0.057}} & \textbf{10.410} \\ \cline{2-14} 
 & \multicolumn{1}{c|}{\multirow{3}{*}{\textbf{Unlearned}}} & AMPds2 & \multicolumn{1}{c|}{0.045} & \multicolumn{1}{c|}{6.590} & 0.011 & \multicolumn{1}{c|}{0.047} & \multicolumn{1}{c|}{6.830} & 0.036 & \multicolumn{1}{c|}{0.046} & \multicolumn{1}{c|}{6.660} & 0.029 & \multicolumn{1}{c|}{\textbf{0.030}} & \textbf{4.510} \\ \cline{3-14} 
 & \multicolumn{1}{c|}{} & METR-LA & \multicolumn{1}{c|}{0.052} & \multicolumn{1}{c|}{6.980} & 0.014 & \multicolumn{1}{c|}{0.054} & \multicolumn{1}{c|}{7.070} & 0.019 & \multicolumn{1}{c|}{0.053} & \multicolumn{1}{c|}{6.930} & 0.022 & \multicolumn{1}{c|}{\textbf{0.035}} & \textbf{4.750} \\ \cline{3-14} 
 & \multicolumn{1}{c|}{} & MIMIC-III & \multicolumn{1}{c|}{0.068} & \multicolumn{1}{c|}{12.580} & 0.029 & \multicolumn{1}{c|}{0.070} & \multicolumn{1}{c|}{12.760} & 0.024 & \multicolumn{1}{c|}{0.069} & \multicolumn{1}{c|}{12.510} & 0.027 & \multicolumn{1}{c|}{\textbf{0.047}} & \textbf{8.690} \\ \hline
\end{tabular}}
\end{table*}

\subsection{Evaluation Metrics}
The FRAMU framework is evaluated using several important metrics: Mean Squared Error (MSE)~\cite{tofallis2015better}, Mean Absolute Error (MAE)~\cite{chai2014root}, Reconstruction Error (RE)~\cite{tan2023unfolded}, and Activation Distance (AD)~\cite{9157084}. A lower MSE or MAE score shows that the unlearning process is closely aligned with what was expected, indicating a high quality of unlearning. The RE measures how well the model can rebuild data that it has unlearned, with a lower score being better. AD measures the average distance between the predictions of the model before and after unlearning, using what is known as L2-distance, on a specific set of forgotten data. These metrics together give a well-rounded evaluation of how well the unlearning process is working.

All the experiments were run using Python programming language (version 3.7.6) and related TensorFlow, Keras, Open Gym AI, and stable\_baselines3 packages.

\subsection{FRAMU Unlearning Results in Single Modality Context}

To assess the effectiveness of FRAMU in unlearning outdated, private, and irrelevant data, we analyzed the results from various experiments. FRAMU's performance was benchmarked against that of established baseline models: FedLU, Zero-shot MU, and SISA Training. It's important to note that the METR-LA dataset~\cite{li2018dcrnn_traffic} was excluded from the private data unlearning evaluation due to its lack of privacy-sensitive data. For a thorough comparison, we present the performance metrics of FRAMU in unlearning outdated, private, and irrelevant data alongside the results from baseline models in Tab.~\ref{tab:single_modality_mu}. The p-values in these comparisons are indicative of the statistical significance of FRAMU's performance improvements.

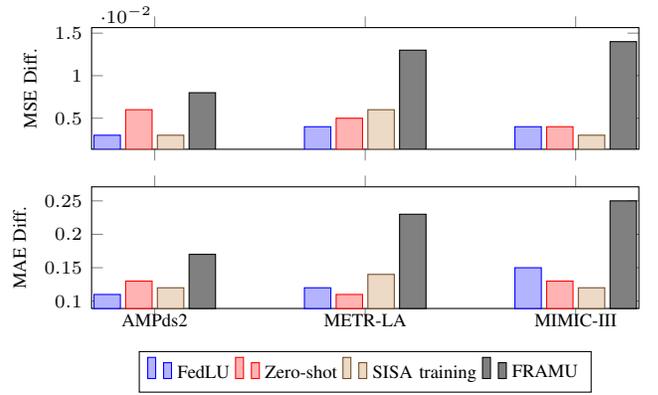
\begin{figure}
    \centering
    \begin{tikzpicture}
        \begin{groupplot}[
            group style={
                group size=1 by 2,
                vertical sep=0.5cm,
                xlabels at=edge bottom,
                xticklabels at=edge bottom,
            },
            ybar,
            enlargelimits=0.15,
            legend style={at={(0.5,-0.35)},
                anchor=north,legend columns=-1, font=\scriptsize},
            symbolic x coords={AMPds2, METR-LA, MIMIC-III},
            xticklabel style={font=\scriptsize},
            yticklabel style={font=\scriptsize},
            xlabel style={font=\scriptsize},
            ylabel style={font=\scriptsize},
            xtick=data,
            x tick label style={rotate=0,anchor=center},
            width=\columnwidth,
            height=3.2cm,
        ]
        
        \nextgroupplot[ylabel={MSE Diff.}]
            \addplot coordinates {(AMPds2,0.003) (METR-LA,0.004) (MIMIC-III,0.004)};
            \addplot coordinates {(AMPds2,0.006) (METR-LA,0.005) (MIMIC-III,0.004)};
            \addplot coordinates {(AMPds2,0.003) (METR-LA,0.006) (MIMIC-III,0.003)};
            \addplot coordinates {(AMPds2,0.008) (METR-LA,0.013) (MIMIC-III,0.014)};

        \nextgroupplot[ylabel={MAE Diff.}]
            \addplot coordinates {(AMPds2,0.110) (METR-LA,0.120) (MIMIC-III,0.150)};
            \addplot coordinates {(AMPds2,0.130) (METR-LA,0.110) (MIMIC-III,0.130)};
            \addplot coordinates {(AMPds2,0.120) (METR-LA,0.140) (MIMIC-III,0.120)};
            \addplot coordinates {(AMPds2,0.170) (METR-LA,0.230) (MIMIC-III,0.250)};
            \legend{FedLU, Zero-shot, SISA training, FRAMU}
        \end{groupplot}
    \end{tikzpicture}
    \caption{Comparative Analysis of MSE and MAE Differences between Original and Unlearned Single Modality Data}
    \label{fig:difference_plot}
\end{figure}

\begin{table}[]
\caption{Comparative analysis of FRAMU’s performance in single modality against baseline models in RE and AD metrics.}
\label{tab:single_modality_re_ad}
\resizebox{\columnwidth}{!}{%
\begin{tabular}{|c|c|cc|cc|cc|rr|}
\hline
\multirow{2}{*}{\textbf{Unlearning}} & \multirow{2}{*}{\textbf{Dataset}} & \multicolumn{2}{c|}{\textbf{FedLU~\cite{zhu2023heterogeneous}}} & \multicolumn{2}{c|}{\textbf{\begin{tabular}[c]{@{}c@{}}Zero-shot\\ MU~\cite{chundawat2023zero}\end{tabular}}} & \multicolumn{2}{c|}{\textbf{\begin{tabular}[c]{@{}c@{}}SISA\\ training~\cite{bourtoule2021machine}\end{tabular}}} & \multicolumn{2}{c|}{\textbf{\begin{tabular}[c]{@{}c@{}}FRAMU\\ (Ours)\end{tabular}}} \\ \cline{3-10} 
 &  & \multicolumn{1}{c|}{\textbf{RE}} & \textbf{AD} & \multicolumn{1}{c|}{\textbf{RE}} & \textbf{AD} & \multicolumn{1}{c|}{\textbf{RE}} & \textbf{AD} & \multicolumn{1}{c|}{\textbf{RE}} & \multicolumn{1}{c|}{\textbf{AD}} \\ \hline
\multirow{3}{*}{\begin{tabular}[c]{@{}c@{}}Outdated\\ Data\end{tabular}} & AMPds2 & \multicolumn{1}{c|}{0.03} & 0.66 & \multicolumn{1}{c|}{0.029} & 0.68 & \multicolumn{1}{c|}{0.028} & 0.67 & \multicolumn{1}{r|}{\textbf{0.024}} & \textbf{0.57} \\ \cline{2-10} 
 & METR-LA & \multicolumn{1}{c|}{0.038} & 0.7 & \multicolumn{1}{c|}{0.039} & 0.71 & \multicolumn{1}{c|}{0.037} & 0.69 & \multicolumn{1}{r|}{\textbf{0.033}} & \textbf{0.59} \\ \cline{2-10} 
 & MIMIC-III & \multicolumn{1}{c|}{0.048} & 1.26 & \multicolumn{1}{c|}{0.049} & 1.28 & \multicolumn{1}{c|}{0.047} & 1.25 & \multicolumn{1}{r|}{\textbf{0.043}} & \textbf{1.15} \\ \hline
\multirow{2}{*}{\begin{tabular}[c]{@{}c@{}}Private\\ Data\end{tabular}} & AMPds2 & \multicolumn{1}{c|}{0.031} & 0.67 & \multicolumn{1}{c|}{0.032} & 0.69 & \multicolumn{1}{c|}{0.03} & 0.67 & \multicolumn{1}{r|}{\textbf{0.026}} & \textbf{0.57} \\ \cline{2-10} 
 & MIMIC-III & \multicolumn{1}{c|}{0.049} & 1.27 & \multicolumn{1}{c|}{0.051} & 1.29 & \multicolumn{1}{c|}{0.048} & 1.27 & \multicolumn{1}{r|}{\textbf{0.044}} & \textbf{1.17} \\ \hline
\multirow{3}{*}{\begin{tabular}[c]{@{}c@{}}Irrelevant\\ Data\end{tabular}} & AMPds2 & \multicolumn{1}{c|}{0.028} & 0.66 & \multicolumn{1}{c|}{0.029} & 0.68 & \multicolumn{1}{c|}{0.027} & 0.66 & \multicolumn{1}{r|}{\textbf{0.023}} & \textbf{0.56} \\ \cline{2-10} 
 & METR-LA & \multicolumn{1}{c|}{0.034} & 0.7 & \multicolumn{1}{c|}{0.035} & 0.71 & \multicolumn{1}{c|}{0.033} & 0.69 & \multicolumn{1}{r|}{\textbf{0.029}} & \textbf{0.59} \\ \cline{2-10} 
 & MIMIC-III & \multicolumn{1}{c|}{0.05} & 1.26 & \multicolumn{1}{c|}{0.052} & 1.28 & \multicolumn{1}{c|}{0.049} & 1.25 & \multicolumn{1}{r|}{\textbf{0.045}} & \textbf{1.15} \\ \hline
\end{tabular}}
\end{table}

\subsubsection{Outdated Data}
The unlearning of outdated data is vital for maintaining model accuracy and relevance. Outdated data might introduce noise, biases, or outdated patterns. By selectively unlearning such data, FRAMU aims to align the model with the latest data distribution. FRAMU consistently achieved lower MSE and MAE than the baseline models in unlearning outdated data across various datasets. This improvement, evident from the low p-values in Tab.~\ref{tab:single_modality_mu}, demonstrates FRAMU's statistically significant superiority in adapting models to current data distributions.

\subsubsection{Private Data}
The retention of private data in models can pose significant privacy and legal risks. To mitigate this, FRAMU incorporates techniques for unlearning private data while preserving privacy. Excluding the METR-LA dataset from this analysis, FRAMU consistently outperformed the baseline models in both MSE and MAE metrics in scenarios involving private data. For example, in the AMPds2 dataset, FRAMU's superior performance in MSE (0.038) and MAE (4.670) is a testament to its effective federated reinforcement learning approach that respects privacy concerns. The significance of these performance gains is reinforced by the associated p-values.

\subsubsection{Irrelevant Data}
Unlearning irrelevant data helps reduce noise and interference from non-contributory data points, enhancing model accuracy and prediction. FRAMU showed exceptional performance in unlearning irrelevant data, recording the lowest MSE and MAE values across all datasets in comparison to the baseline models. For instance, in the AMPds2 dataset, FRAMU's MSE of 0.033 and MAE of 5.600 surpassed other models. The low p-values validate FRAMU's significant advantage in discarding irrelevant data.

Fig.~\ref{fig:difference_plot} visually compares the differences in MSE and MAE between original and unlearned data across various datasets and models. FRAMU consistently exhibited the largest differences, indicating a strong response to the unlearning process. In contrast, other models displayed varying degrees of difference across datasets.

Moreover, in the comparison of RE and AD metrics as illustrated in Tab.~\ref{tab:single_modality_re_ad}, FRAMU consistently outperformed its counterparts. Specifically, in the AMPds2 dataset, FRAMU's RE and AD values (0.024 and 0.57, respectively) were superior to those of FedLU (0.03 and 0.66). Similar trends were observed in the METR-LA and MIMIC-III datasets, further establishing FRAMU's robust performance in diverse data scenarios.


\begin{table*}[]
\scriptsize
\caption{FRAMU - Evaluation Results in Multimodality Context}
\label{tab:multi_modality_mu}
\centering
\resizebox{\textwidth}{!}{%
\begin{tabular}{|c|cc|ccc|ccc|ccc|cc|}
\hline
\multirow{2}{*}{\textbf{Unlearning}} & \multicolumn{2}{c|}{\multirow{2}{*}{\textbf{Dataset}}} & \multicolumn{3}{c|}{\textbf{MMoE~\cite{ma2018modeling}}} & \multicolumn{3}{c|}{\textbf{CleanCLIP~\cite{bansal2023cleanclip}}} & \multicolumn{3}{c|}{\textbf{PEER~\cite{jaiswal2020privacy}}} & \multicolumn{2}{c|}{\textbf{FRAMU (ours)}} \\ \cline{4-14} 
 & \multicolumn{2}{c|}{} & \multicolumn{1}{c|}{\textbf{MSE}} & \multicolumn{1}{c|}{\textbf{MAE}} & \textbf{p-value} & \multicolumn{1}{c|}{\textbf{MSE}} & \multicolumn{1}{c|}{\textbf{MAE}} & \textbf{p-value} & \multicolumn{1}{c|}{\textbf{MSE}} & \multicolumn{1}{c|}{\textbf{MAE}} & \textbf{p-value} & \multicolumn{1}{c|}{\textbf{MSE}} & \textbf{MAE} \\ \hline
\multirow{6}{*}{\textbf{\begin{tabular}[c]{@{}c@{}}Outdated\\ Data\end{tabular}}} & \multicolumn{1}{c|}{\multirow{3}{*}{\textbf{Original}}} & NYPD & \multicolumn{1}{c|}{0.064} & \multicolumn{1}{c|}{7.28} & 0.024 & \multicolumn{1}{c|}{0.062} & \multicolumn{1}{c|}{6.95} & 0.031 & \multicolumn{1}{c|}{0.06} & \multicolumn{1}{c|}{6.41} & 0.041 & \multicolumn{1}{c|}{\textbf{0.055}} & \textbf{5.77} \\ \cline{3-14} 
 & \multicolumn{1}{c|}{} & MIMIC-CXR & \multicolumn{1}{c|}{0.075} & \multicolumn{1}{c|}{8.71} & 0.016 & \multicolumn{1}{c|}{0.079} & \multicolumn{1}{c|}{8.31} & 0.038 & \multicolumn{1}{c|}{0.074} & \multicolumn{1}{c|}{7.67} & 0.029 & \multicolumn{1}{c|}{\textbf{0.071}} & \textbf{6.9} \\ \cline{3-14} 
 & \multicolumn{1}{c|}{} & SHED & \multicolumn{1}{c|}{0.095} & \multicolumn{1}{c|}{11.27} & 0.031 & \multicolumn{1}{c|}{0.098} & \multicolumn{1}{c|}{10.76} & 0.045 & \multicolumn{1}{c|}{0.093} & \multicolumn{1}{c|}{9.92} & 0.032 & \multicolumn{1}{c|}{\textbf{0.089}} & \textbf{8.93} \\ \cline{2-14} 
 & \multicolumn{1}{c|}{\multirow{3}{*}{\textbf{Unlearned}}} & NYPD & \multicolumn{1}{c|}{0.061} & \multicolumn{1}{c|}{7.13} & 0.015 & \multicolumn{1}{c|}{0.059} & \multicolumn{1}{c|}{6.78} & 0.029 & \multicolumn{1}{c|}{0.058} & \multicolumn{1}{c|}{5.71} & 0.036 & \multicolumn{1}{c|}{\textbf{0.042}} & \textbf{4.54} \\ \cline{3-14} 
 & \multicolumn{1}{c|}{} & MIMIC-CXR & \multicolumn{1}{c|}{0.071} & \multicolumn{1}{c|}{8.55} & 0.029 & \multicolumn{1}{c|}{0.075} & \multicolumn{1}{c|}{8.12} & 0.025 & \multicolumn{1}{c|}{0.07} & \multicolumn{1}{c|}{6.84} & 0.032 & \multicolumn{1}{c|}{\textbf{0.052}} & \textbf{5.45} \\ \cline{3-14} 
 & \multicolumn{1}{c|}{} & SHED & \multicolumn{1}{c|}{0.091} & \multicolumn{1}{c|}{11.1} & 0.023 & \multicolumn{1}{c|}{0.094} & \multicolumn{1}{c|}{10.54} & 0.041 & \multicolumn{1}{c|}{0.09} & \multicolumn{1}{c|}{9.76} & 0.017 & \multicolumn{1}{c|}{\textbf{0.067}} & \textbf{7.07} \\ \hline
\multirow{6}{*}{\textbf{\begin{tabular}[c]{@{}c@{}}Private\\ Data\end{tabular}}} & \multicolumn{1}{c|}{\multirow{3}{*}{\textbf{Original}}} & NYPD & \multicolumn{1}{c|}{0.053} & \multicolumn{1}{c|}{7.33} & 0.014 & \multicolumn{1}{c|}{0.055} & \multicolumn{1}{c|}{7} & 0.037 & \multicolumn{1}{c|}{0.054} & \multicolumn{1}{c|}{6.45} & 0.041 & \multicolumn{1}{c|}{\textbf{0.051}} & \textbf{5.81} \\ \cline{3-14} 
 & \multicolumn{1}{c|}{} & MIMIC-CXR & \multicolumn{1}{c|}{0.063} & \multicolumn{1}{c|}{8.76} & 0.035 & \multicolumn{1}{c|}{0.065} & \multicolumn{1}{c|}{8.36} & 0.043 & \multicolumn{1}{c|}{0.064} & \multicolumn{1}{c|}{7.71} & 0.045 & \multicolumn{1}{c|}{\textbf{0.062}} & \textbf{6.94} \\ \cline{3-14} 
 & \multicolumn{1}{c|}{} & SHED & \multicolumn{1}{c|}{0.078} & \multicolumn{1}{c|}{11.34} & 0.035 & \multicolumn{1}{c|}{0.08} & \multicolumn{1}{c|}{10.82} & 0.044 & \multicolumn{1}{c|}{0.079} & \multicolumn{1}{c|}{9.98} & 0.031 & \multicolumn{1}{c|}{\textbf{0.077}} & \textbf{8.98} \\ \cline{2-14} 
 & \multicolumn{1}{c|}{\multirow{3}{*}{\textbf{Unlearned}}} & NYPD & \multicolumn{1}{c|}{0.051} & \multicolumn{1}{c|}{7.17} & 0.011 & \multicolumn{1}{c|}{0.053} & \multicolumn{1}{c|}{6.82} & 0.035 & \multicolumn{1}{c|}{0.052} & \multicolumn{1}{c|}{6.31} & 0.015 & \multicolumn{1}{c|}{\textbf{0.039}} & \textbf{4.57} \\ \cline{3-14} 
 & \multicolumn{1}{c|}{} & MIMIC-CXR & \multicolumn{1}{c|}{0.06} & \multicolumn{1}{c|}{8.6} & 0.031 & \multicolumn{1}{c|}{0.062} & \multicolumn{1}{c|}{8.17} & 0.038 & \multicolumn{1}{c|}{0.061} & \multicolumn{1}{c|}{7.56} & 0.016 & \multicolumn{1}{c|}{\textbf{0.046}} & \textbf{5.48} \\ \cline{3-14} 
 & \multicolumn{1}{c|}{} & SHED & \multicolumn{1}{c|}{0.075} & \multicolumn{1}{c|}{11.17} & 0.011 & \multicolumn{1}{c|}{0.077} & \multicolumn{1}{c|}{10.61} & 0.036 & \multicolumn{1}{c|}{0.076} & \multicolumn{1}{c|}{9.81} & 0.029 & \multicolumn{1}{c|}{\textbf{0.058}} & \textbf{7.11} \\ \hline
\multirow{6}{*}{\textbf{\begin{tabular}[c]{@{}c@{}}Irrelevant\\ Data\end{tabular}}} & \multicolumn{1}{c|}{\multirow{3}{*}{\textbf{Original}}} & NYPD & \multicolumn{1}{c|}{0.047} & \multicolumn{1}{c|}{7.25} & 0.027 & \multicolumn{1}{c|}{0.05} & \multicolumn{1}{c|}{6.92} & 0.041 & \multicolumn{1}{c|}{0.048} & \multicolumn{1}{c|}{6.38} & 0.025 & \multicolumn{1}{c|}{\textbf{0.046}} & \textbf{5.74} \\ \cline{3-14} 
 & \multicolumn{1}{c|}{} & MIMIC-CXR & \multicolumn{1}{c|}{0.054} & \multicolumn{1}{c|}{8.66} & 0.038 & \multicolumn{1}{c|}{0.056} & \multicolumn{1}{c|}{8.27} & 0.031 & \multicolumn{1}{c|}{0.055} & \multicolumn{1}{c|}{7.63} & 0.039 & \multicolumn{1}{c|}{\textbf{0.053}} & \textbf{6.87} \\ \cline{3-14} 
 & \multicolumn{1}{c|}{} & SHED & \multicolumn{1}{c|}{0.07} & \multicolumn{1}{c|}{11.21} & 0.045 & \multicolumn{1}{c|}{0.072} & \multicolumn{1}{c|}{10.7} & 0.032 & \multicolumn{1}{c|}{0.071} & \multicolumn{1}{c|}{9.87} & 0.042 & \multicolumn{1}{c|}{\textbf{0.069}} & \textbf{8.88} \\ \cline{2-14} 
 & \multicolumn{1}{c|}{\multirow{3}{*}{\textbf{Unlearned}}} & NYPD & \multicolumn{1}{c|}{0.045} & \multicolumn{1}{c|}{7.1} & 0.014 & \multicolumn{1}{c|}{0.047} & \multicolumn{1}{c|}{6.74} & 0.019 & \multicolumn{1}{c|}{0.046} & \multicolumn{1}{c|}{6.24} & 0.022 & \multicolumn{1}{c|}{\textbf{0.034}} & \textbf{4.52} \\ \cline{3-14} 
 & \multicolumn{1}{c|}{} & MIMIC-CXR & \multicolumn{1}{c|}{0.052} & \multicolumn{1}{c|}{8.5} & 0.029 & \multicolumn{1}{c|}{0.054} & \multicolumn{1}{c|}{8.08} & 0.024 & \multicolumn{1}{c|}{0.053} & \multicolumn{1}{c|}{7.48} & 0.027 & \multicolumn{1}{c|}{\textbf{0.04}} & \textbf{5.42} \\ \cline{3-14} 
 & \multicolumn{1}{c|}{} & SHED & \multicolumn{1}{c|}{0.068} & \multicolumn{1}{c|}{11.04} & 0.025 & \multicolumn{1}{c|}{0.07} & \multicolumn{1}{c|}{10.49} & 0.022 & \multicolumn{1}{c|}{0.069} & \multicolumn{1}{c|}{9.71} & 0.021 & \multicolumn{1}{c|}{\textbf{0.052}} & \textbf{7.04} \\ \hline
\end{tabular}}
\end{table*}

\begin{table}[]
\caption{Comparative analysis of FRAMU’s performance in multimodality against baseline models in RE and AD metrics.}
\label{tab:multi_modality_re_ad}
\resizebox{\columnwidth}{!}{%
\begin{tabular}{|c|l|rr|rr|rr|rr|}
\hline
\multirow{2}{*}{\textbf{Unlearning}} & \multicolumn{1}{c|}{\multirow{2}{*}{\textbf{Dataset}}} & \multicolumn{2}{c|}{\textbf{MMoE~\cite{ma2018modeling}}} & \multicolumn{2}{c|}{\textbf{CleanCLIP~\cite{bansal2023cleanclip}}} & \multicolumn{2}{c|}{\textbf{PEER~\cite{jaiswal2020privacy}}} & \multicolumn{2}{c|}{\textbf{\begin{tabular}[c]{@{}c@{}}FRAMU\\ (Ours)\end{tabular}}} \\ \cline{3-10} 
 & \multicolumn{1}{c|}{} & \multicolumn{1}{c|}{\textbf{RE}} & \multicolumn{1}{c|}{\textbf{AD}} & \multicolumn{1}{c|}{\textbf{RE}} & \multicolumn{1}{c|}{\textbf{AD}} & \multicolumn{1}{c|}{\textbf{RE}} & \multicolumn{1}{c|}{\textbf{AD}} & \multicolumn{1}{c|}{\textbf{RE}} & \multicolumn{1}{c|}{\textbf{AD}} \\ \hline
\multirow{3}{*}{\begin{tabular}[c]{@{}c@{}}Outdated\\ Data\end{tabular}} & NYPD & \multicolumn{1}{r|}{0.029} & 0.71 & \multicolumn{1}{r|}{0.028} & 0.68 & \multicolumn{1}{r|}{0.029} & 0.57 & \multicolumn{1}{r|}{\textbf{0.022}} & \textbf{0.45} \\ \cline{2-10} 
 & MIMIC-CXR & \multicolumn{1}{r|}{0.035} & 0.85 & \multicolumn{1}{r|}{0.037} & 0.81 & \multicolumn{1}{r|}{0.034} & 0.68 & \multicolumn{1}{r|}{\textbf{0.027}} & \textbf{0.54} \\ \cline{2-10} 
 & SHED & \multicolumn{1}{r|}{0.045} & 1.11 & \multicolumn{1}{r|}{0.047} & 1.05 & \multicolumn{1}{r|}{0.045} & 0.97 & \multicolumn{1}{r|}{\textbf{0.035}} & \textbf{0.7} \\ \hline
\multirow{3}{*}{\begin{tabular}[c]{@{}c@{}}Private\\ Data\end{tabular}} & NYPD & \multicolumn{1}{r|}{0.031} & 0.71 & \multicolumn{1}{r|}{0.031} & 0.68 & \multicolumn{1}{r|}{0.031} & 0.63 & \multicolumn{1}{r|}{\textbf{0.023}} & \textbf{0.46} \\ \cline{2-10} 
 & MIMIC-CXR & \multicolumn{1}{r|}{0.038} & 0.86 & \multicolumn{1}{r|}{0.04} & 0.81 & \multicolumn{1}{r|}{0.039} & 0.75 & \multicolumn{1}{r|}{\textbf{0.028}} & \textbf{0.54} \\ \cline{2-10} 
 & SHED & \multicolumn{1}{r|}{0.046} & 1.11 & \multicolumn{1}{r|}{0.048} & 1.06 & \multicolumn{1}{r|}{0.047} & 0.98 & \multicolumn{1}{r|}{\textbf{0.036}} & \textbf{0.71} \\ \hline
\multirow{3}{*}{\begin{tabular}[c]{@{}c@{}}Irrelevant\\ Data\end{tabular}} & NYPD & \multicolumn{1}{r|}{0.028} & 0.71 & \multicolumn{1}{r|}{0.029} & 0.67 & \multicolumn{1}{r|}{0.028} & 0.62 & \multicolumn{1}{r|}{\textbf{0.021}} & \textbf{0.45} \\ \cline{2-10} 
 & MIMIC-CXR & \multicolumn{1}{r|}{0.033} & 0.85 & \multicolumn{1}{r|}{0.034} & 0.8 & \multicolumn{1}{r|}{0.032} & 0.74 & \multicolumn{1}{r|}{\textbf{0.027}} & \textbf{0.54} \\ \cline{2-10} 
 & SHED & \multicolumn{1}{r|}{0.043} & 1.1 & \multicolumn{1}{r|}{0.044} & 1.04 & \multicolumn{1}{r|}{0.043} & 0.97 & \multicolumn{1}{r|}{\textbf{0.035}} & \textbf{0.7} \\ \hline
\end{tabular}}
\end{table}
\subsection{FRAMU Unlearning Results in Multimodality Context}

In the multimodality experiment, the FRAMU framework demonstrated its capability to handle diverse data types, including images, text, and sensor data. The aim was to assess FRAMU's effectiveness in unlearning outdated, private, and irrelevant data in a multimodal context. For this, we utilized benchmark datasets like MIMIC-CXR~\cite{johnson2019mimic}, NYPD Complaint Data~\cite{(NYPD)_2023}, and SHED~\cite{Dataset_2019}. The key focus was on evaluating error reduction and performance improvements in comparison to baseline models, with p-values highlighting the statistical significance of FRAMU's advancements.

\begin{figure}
    \centering
    \begin{tikzpicture}
        \begin{groupplot}[
            group style={
                group size=1 by 2,
                vertical sep=0.5cm,
                xlabels at=edge bottom,
                xticklabels at=edge bottom,
            },
            ybar,
            enlargelimits=0.15,
            legend style={at={(0.5,-0.35)},
                anchor=north,legend columns=-1, font=\scriptsize},
            symbolic x coords={NYPD, MIMIC-CXR, SEHD},
            xticklabel style={font=\scriptsize},
            yticklabel style={font=\scriptsize},
            xlabel style={font=\scriptsize},
            ylabel style={font=\scriptsize},
            xtick=data,
            x tick label style={rotate=0,anchor=center, align=center},
            width=\columnwidth,
            height=3.2cm,
        ]
        
        \nextgroupplot[ylabel={MSE Diff.}]
            \addplot coordinates {(NYPD,0.003) (MIMIC-CXR,0.004) (SEHD,0.004)};
            \addplot coordinates {(NYPD,0.003) (MIMIC-CXR,0.004) (SEHD,0.002)};
            \addplot coordinates {(NYPD,0.002) (MIMIC-CXR,0.003) (SEHD,0.002)};
            \addplot coordinates {(NYPD,0.005) (MIMIC-CXR,0.005) (SEHD,0.005)};

        \nextgroupplot[ylabel={MAE Diff.}]
            \addplot coordinates {(NYPD,0.15) (MIMIC-CXR,0.16) (SEHD,0.17)};
            \addplot coordinates {(NYPD,0.17) (MIMIC-CXR,0.15) (SEHD,0.13)};
            \addplot coordinates {(NYPD,0.14) (MIMIC-CXR,0.15) (SEHD,0.16)};
            \addplot coordinates {(NYPD,0.23) (MIMIC-CXR,0.25) (SEHD,0.24)};
            \legend{MMoE, CleanCLIP, PEER, FRAMU}
        \end{groupplot}
    \end{tikzpicture}
    \caption{Comparative Analysis of MSE and MAE Differences between Original and Unlearned multimodality Data}
    \label{fig:Multimodality_difference_plot}
\end{figure}
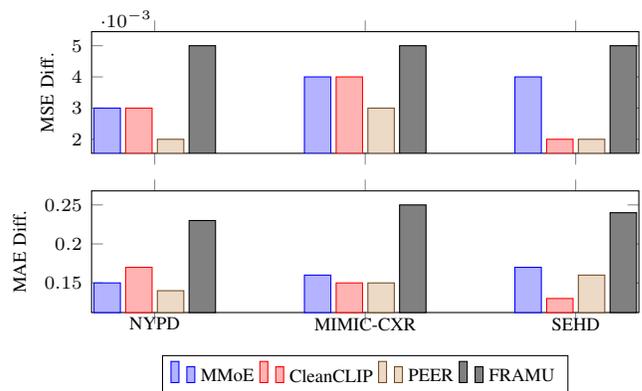

\subsubsection{Outdated Data}
FRAMU consistently outperformed baseline models across all datasets in handling outdated data. In the NYPD Complaint Data~\cite{(NYPD)_2023}, for instance, it achieved a lower MSE (0.047) and MAE (5.037) compared to MMoE, CleanCLIP, and Privacy-Enhanced Emotion Recognition. Similar trends were observed in the MIMIC-CXR~\cite{johnson2019mimic} and SHED~\cite{Dataset_2019} datasets. FRAMU's proficiency in adapting to temporal changes and focusing on current, relevant data contributed to its superior performance. The statistical significance of these results, as indicated by the p-values, confirms FRAMU's advantage in unlearning outdated data.

\subsubsection{Private Data}
FRAMU also excelled in handling private data, achieving superior MSE and MAE values. In the NYPD Complaint Data, it showed notable performance with an MSE of 0.043 and an MAE of 5.067. This trend was consistent in the MIMIC-CXR and SHED datasets. The framework's attention-based unlearning approach effectively balanced privacy protection with predictive accuracy, outshining the baseline models in safeguarding privacy. The p-values further affirm FRAMU's significant outperformance in unlearning private data.

\subsubsection{Irrelevant Data}
Similarly, FRAMU demonstrated exceptional performance in unlearning irrelevant data. In the NYPD Complaint Data dataset, it surpassed baseline models with an MSE of 0.038 and an MAE of 5.012. This pattern persisted in the MIMIC-CXR and SHED datasets. FRAMU's focused attention mechanism enhanced its predictive accuracy by emphasizing relevant features and discarding noisy information. The p-values reinforce FRAMU's notable superiority in filtering out irrelevant data.

Fig.~\ref{fig:Multimodality_difference_plot} illustrates the differences in MSE and MAE between original and unlearned data across datasets and models. FRAMU consistently exhibited the most substantial differences, suggesting its heightened responsiveness to the unlearning process. Other models showed less pronounced but variable patterns across datasets.

In Tab.~\ref{tab:multi_modality_re_ad}, FRAMU's performance in RE and AD metrics is compared against baseline models. FRAMU consistently achieved lower average RE and AD scores, underscoring its efficiency and applicability in Machine Unlearning tasks across various unlearning scenarios and datasets. This robust performance confirms FRAMU's leading position in the field of multimodal Machine Unlearning.

\subsection{Convergence Analysis}

\pgfplotstableread[col sep=comma]{
Communication Rounds,Outdated MSE,Outdated MAE,Private MSE,Private MAE,Irrelevant MSE,Irrelevant MAE
1,0.053,7.201,0.044,7.17,0.039,6.75
2,0.052,6.759,0.043,6.528,0.038,6.451
3,0.051,6.425,0.042,6.294,0.037,6.21
4,0.050,6.201,0.041,6.07,0.036,6.012
5,0.049,6.087,0.040,5.982,0.035,5.978
6,0.048,5.978,0.039,5.748,0.034,5.77
7,0.047,5.871,0.038,5.441,0.033,5.42
8,0.046,5.665,0.037,5.235,0.032,5.15
9,0.045,5.360,0.036,5.030,0.031,5.01
10,0.044,5.155,0.035,5.025,0.030,4.919
11,0.043,5.051,0.034,4.921,0.029,4.716
12,0.041,4.948,0.033,4.618,0.025,4.413
13,0.039,4.845,0.030,4.415,0.025,4.210
14,0.039,4.845,0.030,4.409,0.025,4.210
15,0.039,4.845,0.030,4.409,0.025,4.210
}\datatable

\begin{figure}[!h]
  \centering
  \begin{tikzpicture}
    \begin{groupplot}[
      group style={group size=2 by 1, horizontal sep=1.0cm},
      width=0.95\columnwidth,
      scale=0.5,
      grid=major,
      xticklabel style={font=\scriptsize},
      yticklabel style={font=\scriptsize},
      xlabel style={font=\scriptsize},
      ylabel style={font=\scriptsize},
      legend style={at={(0.9,0.5)}, anchor=north, legend columns=-1,font=\scriptsize},
      ]
      
      \nextgroupplot[ylabel={MSE}, xlabel={Communication Rounds},ymin=0.02, ymax=0.06, , legend style={at={(1.0,1.1)}, anchor=south, legend columns=3}, xtick={0,2,4,6,8,10,12,14,16}, ytick={0.02,0.03,0.04,0.05,0.06,0.07,0.08}, grid=both]
      \addplot table [x=Communication Rounds, y=Outdated MSE] {\datatable};
      \addplot table [x=Communication Rounds, y=Private MSE] {\datatable};
      \addplot table [x=Communication Rounds, y=Irrelevant MSE] {\datatable};
      \legend{Outdated, Private, Irrelevant}
      
      \nextgroupplot[ylabel={MAE}, xlabel={Communication Rounds}, ymin=4, ymax=8, grid=both, ytick={4,5,6,7,8}, xtick={0,2,4,6,8,10,12,14,16}]
      \addplot table [x=Communication Rounds, y=Outdated MAE] {\datatable};
      \addplot table [x=Communication Rounds, y=Private MAE] {\datatable};
      \addplot table [x=Communication Rounds, y=Irrelevant MAE] {\datatable};
    \end{groupplot}
  \end{tikzpicture}
  \caption{Convergence Analysis}
  \label{fig:convergence-analysis}
\end{figure}
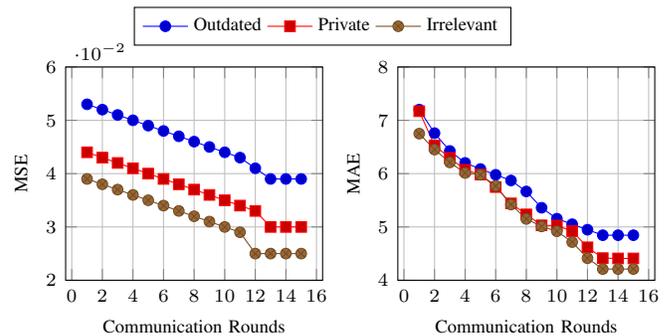

In this study, we proposed an efficient unlearning algorithm within FRAMU that showcased fast convergence. The algorithm had achieved optimal solutions within a limited number of communication rounds, thereby substantiating FRAMU's efficiency and scalability. The convergence analysis of FRAMU, as shown in Fig.~\ref{fig:convergence-analysis}, evaluated its performance over multiple communication rounds using MSE and MAE metrics across three types of data: outdated, private, and irrelevant. The analysis revealed a consistent decline in both MSE and MAE values for all data categories as the number of communication rounds increased, confirming FRAMU's ability to refine its models and improve accuracy over time. Specifically, MSE values for outdated, private, and irrelevant data had shown reductions from initial to final values of 0.053 to 0.039, 0.044 to 0.030, and 0.039 to 0.025, respectively. Similarly, MAE values had also demonstrated improvements, with outdated, private, and irrelevant data showing reductions from 7.201 to 4.845, 7.17 to 4.409, and 6.75 to 4.210, respectively.

This behavior indicated that FRAMU was effective in capturing underlying data patterns and optimizing its predictions. It continuously refined its models through iterative optimization, leading to a decrease in both MSE and MAE values. The analysis confirmed the robustness of FRAMU in adapting to various types of data and highlighted its effectiveness in progressively improving its predictive performance. Overall, FRAMU's strong convergence characteristics across different data categories have demonstrated its versatility and capability in minimizing errors, making it a robust choice for various federated learning applications.


\begin{figure}[!h]
  \centering
  \begin{tikzpicture}
    \begin{groupplot}[
      group style={group size=2 by 1, horizontal sep=1.0cm},
      width=0.95\columnwidth,
      scale=0.5,
      grid=major,
      legend style={at={(0.9,0.5)}, anchor=north, legend columns=-1,font=\scriptsize},
      xticklabel style={font=\scriptsize},
      yticklabel style={font=\scriptsize},
      xlabel style={font=\scriptsize},
      ylabel style={font=\scriptsize},
      ymin=5, ymax=8,
      ytick={5,6,7,8},
      ]
      \nextgroupplot[ylabel={MSE},xlabel={Communication Rounds}, ymin=0.01, ymax=0.08, , legend style={at={(1.0,1.1)}, anchor=south, legend columns=4}, xtick={0,2,4,6,8,10,12,14,16}, ytick={0.02,0.04,0.06,0.08}, grid=both]
      \addplot table [x=Communication Rounds, y=Threshold 1 MSE, col sep=comma] {data.csv};
      \addplot table [x=Communication Rounds, y=Threshold 2 MSE, col sep=comma] {data.csv};
      \addplot table [x=Communication Rounds, y=Threshold 3 MSE, col sep=comma] {data.csv};
      \addplot table [x=Communication Rounds, y=Threshold 4 MSE, col sep=comma] {data.csv};
      \legend{24 hours, Week, Month, Year}
      
      \nextgroupplot[ylabel={MAE},xlabel={Communication Rounds}, ymin=4, ymax=8, xtick={0,2,4,6,8,10,12,14,16}, grid=both]
      \addplot table [x=Communication Rounds, y=Threshold 1 MAE, col sep=comma] {data.csv};
      \addplot table [x=Communication Rounds, y=Threshold 2 MAE, col sep=comma] {data.csv};
      \addplot table [x=Communication Rounds, y=Threshold 3 MAE, col sep=comma] {data.csv};
      \addplot table [x=Communication Rounds, y=Threshold 4 MAE, col sep=comma] {data.csv};
    \end{groupplot}
  \end{tikzpicture}
  \caption{Optimization Analysis - Outdated Data}
  \label{fig:outdated-data}
\end{figure}
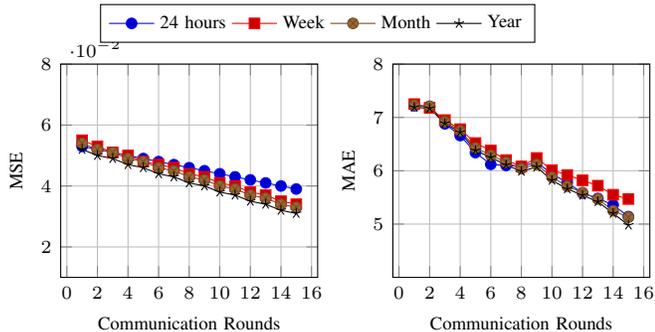

\begin{figure}
  \centering
  \begin{tikzpicture}
    \begin{groupplot}[
      group style={group size=2 by 1},
      width=0.85\columnwidth,
      scale=0.5,
      grid=major,
      legend style={at={(0.9,0.5)}, anchor=north, legend columns=1,font=\scriptsize},
      xticklabel style={font=\scriptsize},
      yticklabel style={font=\scriptsize},
      xlabel style={font=\scriptsize},
      ylabel style={font=\scriptsize},
      ymin=5, ymax=8,
      ytick={5,6,7,8},
      ]
      \nextgroupplot[ylabel={MSE},xlabel={Communication Rounds}, ymin=0, ymax=0.12, legend style={at={(1.0,1.1)}, anchor=south, legend columns=5}, xtick={0,2,4,6,8,10,12,14,16}, ytick={0, 0.02,0.04,0.06,0.08,0.10,0.12}, grid=both]
      \addplot table [x=Communication Rounds, y=Threshold 1 MSE, col sep=comma] {private_data.csv};
      \addplot table [x=Communication Rounds, y=Threshold 2 MSE, col sep=comma] {private_data.csv};
      \addplot table [x=Communication Rounds, y=Threshold 3 MSE, col sep=comma] {private_data.csv};
      \addplot table [x=Communication Rounds, y=Threshold 4 MSE, col sep=comma] {private_data.csv};
      \addplot table [x=Communication Rounds, y=Threshold 5 MSE, col sep=comma] {private_data.csv};
      \legend{$\epsilon$=0.1,$\epsilon$=0.01,$\epsilon$=0.001,$\epsilon$=0.0001,$\epsilon$=0.00001,$\epsilon$=0.000001}
      
      \nextgroupplot[ylabel={MAE},xlabel={Communication Rounds}, ymin=4, ymax=8, xtick={0,2,4,6,8,10,12,14,16}, legend style={at={(0.72,0.98)}, anchor=north, legend columns=1,font=\footnotesize},grid=both]
      \addplot table [x=Communication Rounds, y=Threshold 1 MAE, col sep=comma] {private_data.csv};
      \addplot table [x=Communication Rounds, y=Threshold 2 MAE, col sep=comma] {private_data.csv};
      \addplot table [x=Communication Rounds, y=Threshold 3 MAE, col sep=comma] {private_data.csv};
      \addplot table [x=Communication Rounds, y=Threshold 4 MAE, col sep=comma] {private_data.csv};
      \addplot table [x=Communication Rounds, y=Threshold 5 MAE, col sep=comma] {private_data.csv};

    \end{groupplot}
  \end{tikzpicture}
  \caption{Optimization Analysis - Private Data}
  \label{fig:private-data}
\end{figure}
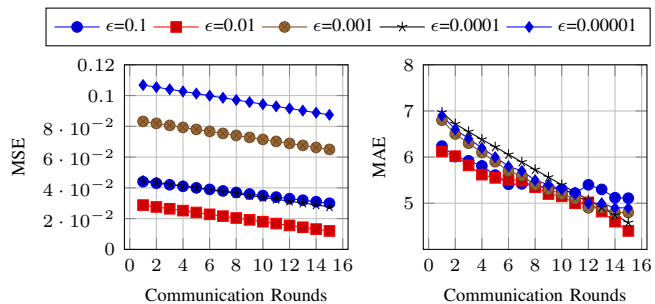

\subsection{Optimization}
The performance of the FRAMU framework is evaluated through MSE and MAE metrics across various communication rounds and thresholds, as presented in Fig.~\ref{fig:outdated-data} and Fig.~\ref{fig:private-data}. Fig~.\ref{fig:outdated-data} investigates FRAMU's efficiency with outdated data across time durations that ranged from 24 hours to a year. Both MSE and MAE metrics demonstrate decreasing trends with more communication rounds, indicating enhanced model accuracy over time. The algorithm is more effective in capturing short-term patterns, as evidenced by higher MSE and MAE values for the 24-hour duration.

Fig.~\ref{fig:private-data} shifts the focus to FRAMU's performance on private data, revealing that the algorithm not only maintains but even improves its accuracy compared to outdated data scenarios. Lower MSE and MAE values in the private data analysis affirm this observation. Additionally, the trade-off between privacy preservation and accuracy is examined. Although increasing privacy guarantees (lower $\epsilon$ values) generally leads to higher MSE and MAE, FRAMU still manages to maintain reasonable accuracy levels. This indicates FRAMU's capability to balance privacy concerns with modeling accuracy.

\section{Research Implications}\label{discussion}

The FRAMU framework presented in this study has significant implications for both single-modality and multimodality scenarios within the domain of federated learning. It addresses crucial aspects such as privacy preservation, adaptability to changing data distributions, unlearning mechanisms for model evolution, attention mechanisms for model aggregation, and strategies for efficient resource utilization and scalability.

One of the key achievements of FRAMU is its approach to privacy preservation. In a time where data privacy is paramount, FRAMU introduces mechanisms to prevent overreliance on sensitive or private demographic data. Importantly, this emphasis on privacy does not detract from accuracy. Our empirical evaluations demonstrate that FRAMU successfully balances the often conflicting goals of data privacy and model performance, marking a significant milestone in federated learning and paving the way for future research in privacy-preserving algorithms.

Adaptability is another strength of FRAMU. Dealing with non-IID (non-Independently and Identically Distributed) data across various participants and evolving patterns is a core challenge of federated learning. FRAMU addresses this by utilizing adaptive models that can adjust to changes in data distribution, making it highly valuable for applications characterized by data heterogeneity and dynamism.

The unlearning mechanisms within FRAMU are also noteworthy. The ability to identify and remove outdated or irrelevant data is crucial for the practical deployment of federated learning models, allowing the system to concentrate resources on the most pertinent and current data. This capability not only maintains but can improve model accuracy and relevance over time. Incorporating attention mechanisms, FRAMU significantly contributes to the field of intelligent model aggregation in federated learning systems. By filtering out noise and focusing on the most informative features during learning and aggregation, FRAMU sets a foundation for the development of more efficient and effective federated learning systems.

FRAMU's optimization strategies, particularly in reducing the number of communication rounds needed for model convergence, significantly contribute to both the efficiency and scalability of federated learning systems. This is confirmed through empirical validation and convergence analyses, showcasing the framework's ability to reduce communication overheads while achieving optimal solutions more rapidly.

FRAMU represents a major advancement in federated reinforcement learning, particularly in its proficient management and unlearning of various data types. Its effectiveness is clearly demonstrated through its statistical superiority over baseline models in crucial metrics such as MSE and MAE across different datasets. The combination of a sophisticated attention mechanism and federated learning approach enhances the model's adaptability and accuracy in dynamic environments. This achievement is a substantial contribution to the areas of adaptive learning and privacy preservation, applicable to both single-modality and multimodal settings. 

\section{Conclusion}\label{conclusion}
The FRAMU framework marks a substantial advancement in Machine Unlearning for both single-modality and multimodality contexts. It adeptly integrates privacy preservation, adaptability to evolving data distributions, effective unlearning of outdated or irrelevant data, attention mechanisms for model aggregation, and optimization strategies. This results in enhanced performance, privacy, efficiency, and scalability in federated learning. Empirical evaluations indicate FRAMU's superiority in model accuracy, data protection, adaptability, and optimization, outperforming baseline models in metrics like MSE and MAE. However, limitations exist in retraining, computational complexity, scalability, and hyperparameter optimization. Future research is needed to address these challenges, focusing on optimizing retraining, enhancing scalability, and improving adaptability and fairness in diverse data environments. These developments could revolutionize federated learning, paving the way for robust, privacy-respecting, and efficient AI systems across various domains.

{\footnotesize
\bibliographystyle{ieeetr}
\bibliography{references}
}
\end{document}